\documentclass[11pt]{article}
\usepackage{acl}
\usepackage{times}
\usepackage{latexsym}
\usepackage[T1]{fontenc}
\usepackage[utf8]{inputenc}
\usepackage{microtype}
\usepackage{inconsolata}
\usepackage{graphicx}
\usepackage{amsmath}
\usepackage{amssymb}
\usepackage{booktabs}
 \usepackage{multirow}
\usepackage{enumitem}
\usepackage{pgfplots}
\usepackage{algorithm}
\usepackage{algpseudocode}
\usepackage[most]{tcolorbox}
\pgfplotsset{compat=1.17}
\usepackage{makecell}
\usepackage{array}
\usepackage{colortbl}

\definecolor{lightgray}{gray}{0.9}

\definecolor{missinggray}{gray}{0.55}
\definecolor{deltagray}{gray}{0.38}
\definecolor{gain}{RGB}{28,118,82}
\definecolor{loss}{RGB}{165,72,72}
\definecolor{revrow}{RGB}{239,240,255}
\definecolor{alfours}{RGB}{88,142,126}
\definecolor{alfbase}{RGB}{180,193,202}

\usepackage{float}
\usepackage{placeins}
\newcommand{\missingcell}{\multicolumn{1}{c}{\textcolor{missinggray}{\textemdash}}}

\newcommand{\succcell}[1]{\makebox[3.0em][r]{#1}}
\newcommand{\bestsucc}[1]{\makebox[3.0em][r]{\textbf{#1}}}
\newcommand{\overallcell}[1]{\makebox[4.4em][r]{#1}}
\newcommand{\bestoverall}[1]{\makebox[4.4em][r]{\textbf{#1}}}
\newcommand{\basedelta}{\makebox[3.2em][r]{\textcolor{missinggray}{\textemdash}}}
\newcommand{\deltapp}[1]{\makebox[3.2em][r]{#1}}
\newcommand{\revsh}[1]{\cellcolor{revrow}#1}
\newcommand{\gainpp}[1]{\textcolor{gain}{+#1}}
\newcommand{\losspp}[1]{\textcolor{loss}{#1}}
\newcommand{\zeropp}{\textcolor{deltagray}{+0.00}}

\newcommand{\caseheading}[1]{\par\smallskip\noindent\textbf{#1}\par\smallskip\noindent}
\newcommand{\setupheading}[1]{\par\vspace{3pt}\noindent\textbf{#1}}
\newcommand{\principlecard}[4]{%
  \textbf{#1}\par
  \emph{Trigger:} #2\par
  \emph{Repair:} #3\par
  \emph{Guard:} #4%
}

\definecolor{promptframe}{RGB}{130,160,145}
\definecolor{promptback}{RGB}{247,249,248}
\definecolor{authorframe}{RGB}{88,142,126}
\definecolor{execframe}{RGB}{83,124,165}
\definecolor{diagframe}{RGB}{139,105,157}
\definecolor{revisionframe}{RGB}{145,110,70}
\definecolor{ablationframe}{RGB}{150,90,90}
\definecolor{schemaframe}{RGB}{105,105,105}
\newtcolorbox{promptbox}[2][promptframe]{
  enhanced jigsaw,
  breakable,
  colback=promptback,
  colframe=#1,
  colbacktitle=#1,
  coltitle=white,
  fonttitle=\bfseries\footnotesize,
  title={#2},
  boxrule=0.8pt,
  arc=2mm,
  left=5pt,
  right=5pt,
  top=4pt,
  bottom=4pt,
  before skip=2pt,
  after skip=2pt,
  pad at break*=1pt,
  fontupper=\scriptsize,
  before upper={\raggedright\setlength{\parindent}{0pt}}
}

\title{\textsc{SkillRevise}: Improving LLM-Authored Agent Skills via Trace-Conditioned Skill Revision}
\author{
\textbf{Yuxuan Liu$^{1}$ \quad
Zhaochen Su$^{1}$ \quad
Lingyun Xie$^{2}$ \quad
Yuhao Zhang$^{3}$ \quad
Qing Zong$^{1}$} \\
\textbf{Jiahe Guo$^{2}$ \quad
Zhongwei Xie$^{1}$ \quad
Yiyan Ji$^{4}$ \quad
Yauwai Yim$^{1}$ \quad
Hongyu Luo$^{1}$} \\
\textbf{Xiyu Ren$^{1}$ \quad
Ruan Chenyu$^{5}$ \quad
Haoran Li$^{1}$\textsuperscript{*} \quad
Yangqiu Song$^{1}$} \\
$^{1}$The Hong Kong University of Science and Technology, Hong Kong SAR, China \\
$^{2}$Harbin Institute of Technology, Harbin, China \quad
$^{3}$Harbin Institute of Technology, Shenzhen, China \\
$^{4}$Nanjing University, Nanjing, China \quad
$^{5}$The University of Hong Kong, Hong Kong SAR, China \\
\texttt{\{yliurk,zsubf,hlibt\}@connect.ust.hk, yqsong@cse.ust.hk} \quad
\textsuperscript{*}Corresponding author
}

\begin{document}
\raggedbottom
\maketitle

\begin{abstract}
Agent skills are procedural artifacts that enable LLM agents to execute workflows, verify constraints, and recover from failures. Existing self-evolving methods refine skills using accumulated trajectories. However, they struggle in cold-start settings, where only an initial, imperfect skill is available. Consequently, skill construction defaults to expert authoring or one-shot LLM generation. Expert-authored skills are costly and may not align with how LLM agents actually execute tasks, while one-shot generated skills can be syntactically well formed yet behaviorally weak. To bridge this gap, we propose \textsc{SkillRevise}, an execution-grounded framework designed to iteratively refine these initial skills. \textsc{SkillRevise} diagnoses skill defects from execution evidence, retrieves relevant repair principles from a general memory, and applies execution-anchored edits. By re-executing candidates and measuring empirical utility, it retains the best observed skill within the revision budget. Evaluated across three main benchmarks, two domain-specific studies, and six LLMs, \textsc{SkillRevise} substantially outperforms one-shot baselines---improving the base agent's success rate on SkillsBench from 36.05\% to 61.63\%. Furthermore, the revised skills transfer across both executors and task environments, suggesting that \textsc{SkillRevise} captures reusable procedural knowledge beyond any single executor. Our code is available at \url{https://github.com/HKUST-KnowComp/skillrevise}.

\end{abstract}

\section{Introduction}

\begin{figure}[t]
\centering
\vspace{0.08in}
\makebox[\columnwidth][c]{\includegraphics[width=1.0\columnwidth,trim=0 105 0 105,clip]{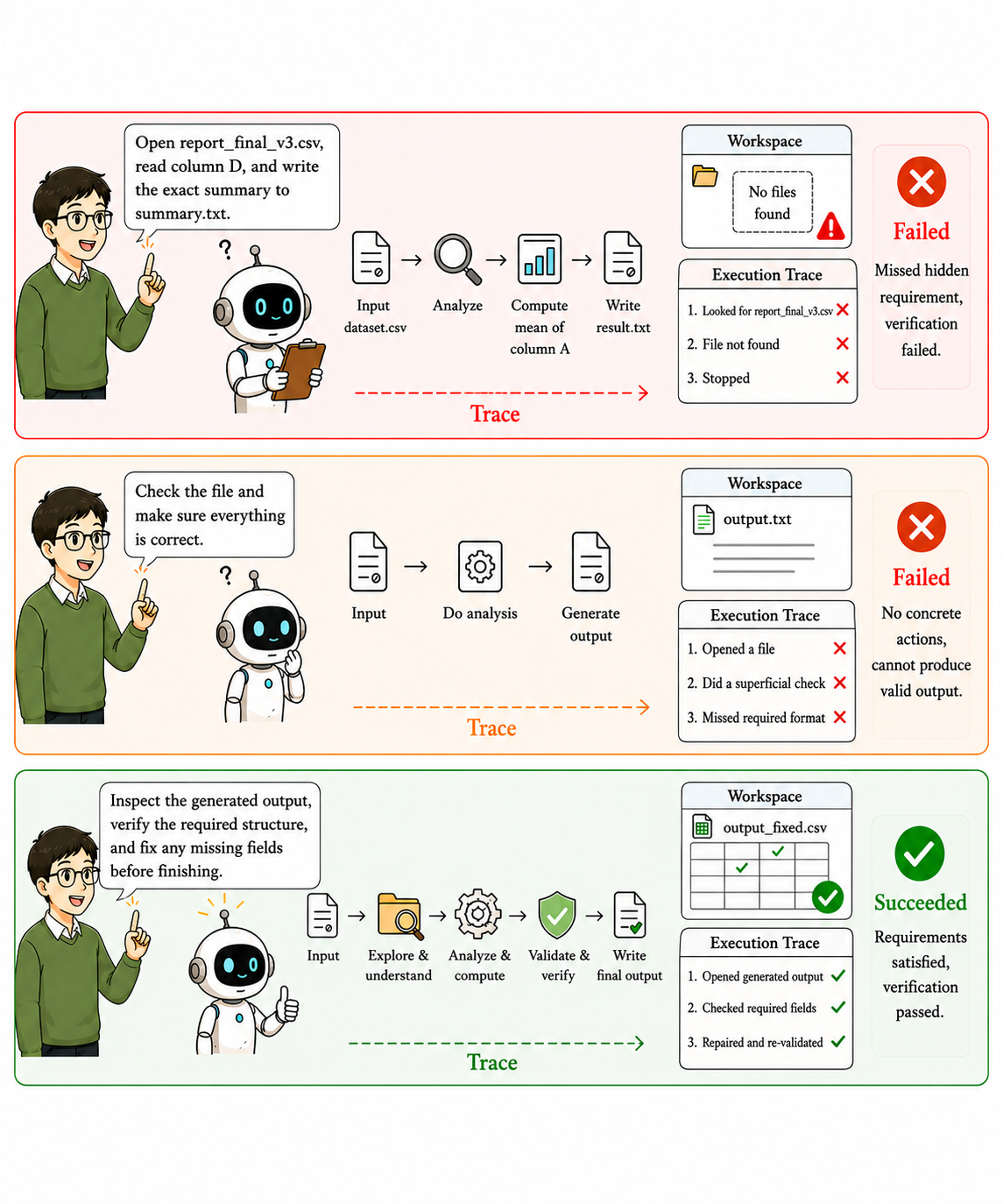}}
\caption{Skill design must avoid both instance-specific shortcuts and vague advice; trace-conditioned principles provide actionable, verifier-aligned guidance.}
\label{fig:skill-design-balance}
\end{figure}

LLM agents are increasingly expected to operate in complex, verifier-driven environments rather than merely produce natural-language answers \citep{li2026skillsbench}. In such settings, success requires agents to inspect workspaces, invoke tools, generate artifacts, satisfy task-specific constraints, and recover from failures \citep{zhou2026skillsurvey}. Existing tools and prompts provide useful capabilities and high-level guidance, but they do not fully specify when to act, how to sequence operations, how to validate outputs, or how to handle errors. \emph{Agent skills} address this gap by externalizing reusable procedural knowledge: unlike atomic tools or one-off prompts, skills organize multi-step workflows, execution constraints, verification checkpoints, and recovery strategies \citep{anthropic2025agentskills}. Figure~\ref{fig:skill-design-balance} illustrates the central design tension: overly instance-specific skills can encode brittle shortcuts, while overly generic skills fail to guide concrete execution; useful skills should instead abstract execution traces into actionable, verifier-aligned principles.

However, skills do not reliably improve agent performance \citep{li2026skillsbench,han2026sweskillsbench}. Tools typically expose well-defined operations with clear input-output behavior. A skill, by contrast, guides how an agent should organize its behavior; its value depends not only on the written instruction itself, but also on how the skill is selected, triggered, executed, and maintained in the target environment \citep{liu2026wildskills}. Recent studies on agent skills show that low-quality or poorly matched skills often yield little improvement \citep{liu2026wildskills} and may even degrade agent performance \citep{li2026skillsbench,han2026sweskillsbench}. Excessive skill content can also dilute attention, motivating token-efficient skill design \citep{gao2026skillreducer}. These findings shift the focus from skill adoption to skill acquisition.

Existing approaches to skill acquisition can be broadly grouped into three categories: retrieval, self-evolution, and direct authoring \citep{zhou2026skillsurvey}. 
Retrieval-based methods reuse skills from existing repositories or memory stores.
However, the selected skills may not fully match the target task and result in disrupted execution and poor adaptivity \citep{liu2026wildskills,han2026sweskillsbench}.
Self-evolution methods refine skills from an agent's own trajectories, failures, feedback, and past behavior \citep{wang2026skillx,ma2026skillclaw}. 
However, this reliance on accumulated experience leaves a cold-start gap when few trajectories are available \citep{zhou2026skillsurvey}.
Direct authoring or generation creates skills from expert knowledge or task-conditioned LLM outputs, which is useful when no appropriate skill is available \citep{anthropic2025agentskills,li2026skillsbench}.
Nevertheless, expert authoring is costly, and one-shot generation may be unreliable without execution-based validation \citep{li2026skillsbench}.

To bridge this gap, we propose \textsc{SkillRevise}, a revision framework built around a task-specific/general decomposition. 
Inspired by agent-learning methods that distill task trajectories into reusable skills \citep{xia2026skillrl} and retain both past experiences and extracted insights \citep{zhao2024expel}, we treat revision as coupling task-specific execution evidence with general repair knowledge. Diagnosis identifies what failed and what should be preserved in the current episode; retrieved repair principles capture recurring design defects. Starting from an LLM-authored skill, \textsc{SkillRevise} executes, diagnoses, revises, re-executes, and retains the best observed version under measured utility within a bounded budget. Thus, it keeps the feedback advantage of self-evolution without requiring a large skill corpus or many prior trajectories.

We evaluate \textsc{SkillRevise} in a unified verifier-driven harness across both standard and out-of-distribution skill-use settings. On SkillsBench \citep{li2026skillsbench}, \textsc{SkillRevise} improves GPT-5.5~\citep{openai2026gpt55} from 36.05\% success without skills and 39.53\% with one-shot skill generation to 61.63\% within a three-round revision budget. To test whether these gains extend beyond the original benchmark distribution, we further adapt SkillLearnBench-Random~\citep{zhong2026skilllearnbench} and SWE-Skills-Bench-Hard~\citep{han2026sweskillsbench} into the same harness, and include calibration-then-freeze principle-absorption studies in ALFWorld~\citep{shridhar2021alfworld} for interactive embodied tasks and BFCL-v4~\citep{patil2025bfcl} for single- and multi-turn function calling. Across these benchmarks and multiple strong executors, including GPT-5.5, Claude Opus 4.7~\citep{anthropic2026opus47}, Kimi K2.6~\citep{moonshotai2026kimik26}, Qwen3.6-Plus~\citep{alibabacloud2026qwen36plus}, and DeepSeek-V4-Pro~\citep{deepseekai2026deepseekv4pro}, execution-grounded revision consistently improves skill performance, suggesting that \textsc{SkillRevise} offers a robust and general approach to automated skill authoring.

This paper makes three contributions:
\begin{itemize}[leftmargin=*,itemsep=2pt]
    \item We formulate cold-start skill improvement as bounded, execution-grounded revision of an existing LLM-authored skill, rather than retrieval or open-ended self-evolution from prior trajectories.
    \item We introduce \textsc{SkillRevise}, which combines task-specific Diagnosis, reusable Principle Memory, and execution-anchored revision with utility-gated selection of the best observed skill.
    \item We evaluate \textsc{SkillRevise} across SkillsBench, SkillLearnBench-Random, SWE-Skills-Bench-Hard, and domain-specific ALFWorld and BFCL-v4 studies, showing consistent gains over no-skill and one-shot Skill-Creator baselines across multiple executor models.
\end{itemize}

\section{Related Work}

\noindent\textbf{Agent skills and skill benchmarks.}
Agent skills have emerged as reusable procedural artifacts for extending LLM agents beyond one-off prompts and atomic tools \citep{anthropic2025agentskills,zhou2026skillsurvey,liu2026wildskills,zhong2026skilllearnbench,chen2026skillcraft}. SkillsBench evaluates whether skills improve agents through diverse professional tasks and shows that curated skills can help whereas self-generated skills are unreliable \citep{li2026skillsbench}. SkillLearnBench studies continual learning methods for agent skill generation on real-world tasks, emphasizing whether agents can accumulate task experience into reusable skills over time \citep{zhong2026skilllearnbench}. WildSkills studies more realistic retrieval settings in which agents must select from large skill collections \citep{liu2026wildskills}, and SWE-Skills-Bench evaluates skill injection in software-engineering repositories with deterministic tests \citep{han2026sweskillsbench}. These studies motivate our setting: the value of a skill depends on whether it is aligned with the executor, task context, and verifier, not merely on whether the skill exists. However, these benchmarks mainly evaluate whether skills help or can be learned from experience, rather than how an imperfect initial skill can be revised under bounded execution feedback.

\noindent\textbf{Self-evolving skills.} 
Recent systems learn or evolve reusable skills from experience \citep{liu2026rethinking}. SkillX constructs hierarchical skill knowledge bases from trajectories \citep{wang2026skillx}; SkillRL builds a hierarchical SkillBank and retrieves general and task-specific heuristics for policy improvement \citep{xia2026skillrl}; Skill-Pro learns executable procedural skills using a Skill-MDP and non-parametric PPO \citep{mi2026skillpro}; MemSkill evolves memory operations as skills \citep{zhang2026memskill}; AutoSkill abstracts personalized skills from interaction histories \citep{yang2026autoskill}; EvoSkill discovers and refines reusable skills through failure analysis \citep{alzubi2026evoskill}; CoEvoSkills co-evolves a Skill Generator and Surrogate Verifier to refine structured skill packages without access to ground-truth test content \citep{zhang2026coevoskills}; and SkillClaw evolves skills collectively from multi-user trajectories \citep{ma2026skillclaw}. However, these methods often require accumulated trajectories, repeated practice, or evolving skill libraries, making them less suitable for bounded cold-start revision.

\noindent\textbf{Agent memory, reflection, and verifier-guided improvement.}
Our work is also related to agent memory and self-improvement methods that store experience for future use. Prior systems use episodic, semantic, or dynamically linked memories, retrieved reflections, or accumulated experience-skill knowledge to improve long-horizon interaction and planning \citep{packer2023memgpt,zhong2024memorybank,xu2025amem,jiang2026xskill}. Reflection-based methods use linguistic feedback to guide subsequent attempts \citep{shinn2023reflexion}, while self-improving agents may recursively modify their logic under environmental feedback \citep{yin2025godel}. Software-engineering agents similarly use environmental feedback, including test outcomes, together with step-level reflection during iterative repository-level issue solving \citep{xia2025live}. \textsc{SkillRevise} instead revises a reusable skill artifact and validates its downstream effect under a fixed verifier, rather than only improving a single trajectory or final answer. 

\begin{figure*}[!t]
\centering
\includegraphics[width=0.98\textwidth]{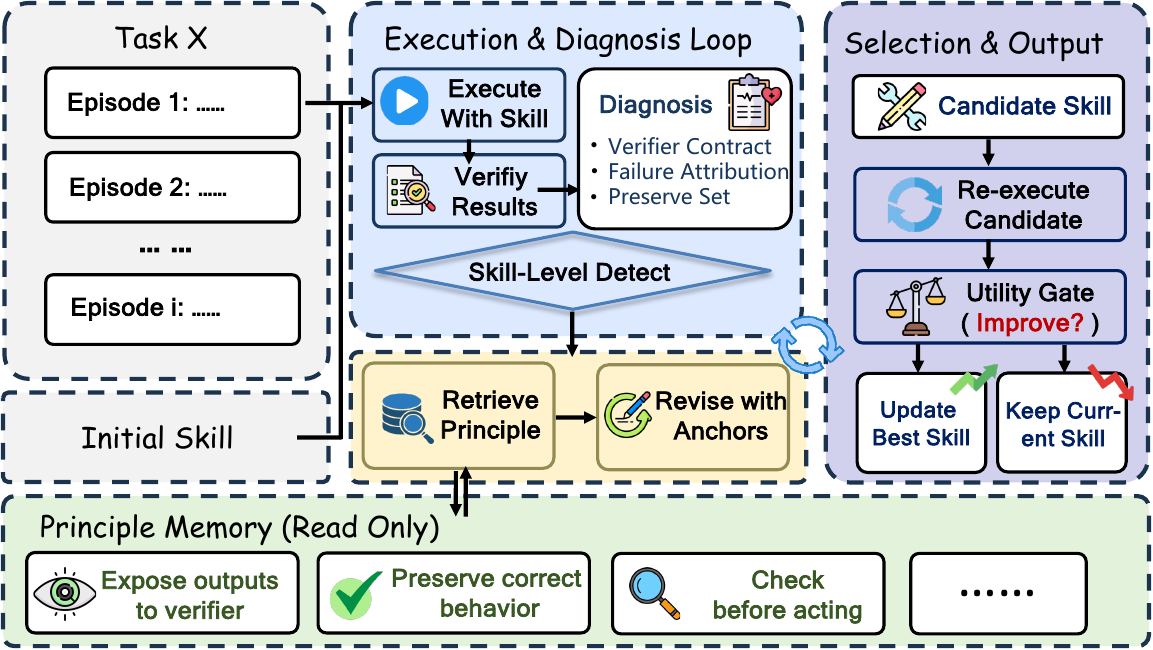}
\caption{\textsc{SkillRevise} pipeline. Within a bounded episode, the solid loop executes and verifies the current skill, diagnoses defects and preservation constraints, retrieves principles from read-only memory, generates and re-executes an anchored candidate, and updates the best observed skill only when utility improves, otherwise retaining the incumbent. The dashed arrow denotes optional post-evaluation memory absorption.}
\vspace{-0.05in}
\label{fig:skillrevise-pipeline}
\end{figure*}

\section{Method: \textsc{SkillRevise}}
\label{sec:method}
This section describes \textsc{SkillRevise} as a bounded, execution-grounded revision loop for improving an existing LLM-authored skill. 
As shown in Figure~\ref{fig:skillrevise-pipeline}, starting from an initial skill $S_0$, the framework executes the skill, diagnoses verifier-facing failures, retrieves and binds repair principles, edits with executable anchors, re-executes the candidate, and applies utility-gated selection. 
We first define the two sources of revision information, Diagnosis and Principle Memory, then formalize the Revision Operator that converts them into concrete skill edits and the bounded episode that governs re-execution and best-observed selection.

\subsection{Diagnosis}

Diagnosis is the task-specific component of \textsc{SkillRevise}. 
For the $i$-th revision attempt, we represent Diagnosis as
\begin{equation}
D_i=(V_i,A_i,\mathcal{K}_i),
\end{equation}
where $V_i$ is the \emph{verification specification}, 
$A_i$ is the \emph{failure attribution}, 
and $\mathcal{K}_i$ is the \emph{preservation constraints}. 
The verification specification $V_i$ describes the observable requirements used to judge the run, including output paths, schemas, formats, terminal sentinels, and pass/fail assertions. 
The attribution component $A_i$ summarizes failed checks, observed behavior, probable causes, and defect labels, indicating whether and how the current skill contributed to failure. 
The preservation constraints $\mathcal{K}_i$ record checks already satisfied and choices likely responsible for them. 
Together, $D_i$ turns raw execution evidence into repair constraints: it specifies what to repair, what evidence supports the repair, and what behavior must remain intact.

\subsection{Principle Memory}

Principle Memory $\mathcal{M}$ stores reusable repair principles rather than task solutions. 
Each principle abstracts a recurring skill-design defect into an operational repair pattern. 
A principle entry specifies when the repair should be considered, what defect it addresses, how the skill should change, what executor action the edit should induce, how the repair should be verified, and when the principle should not transfer. 
The seven seed principles used in the main experiments were manually consolidated from recurring, domain-general skill-design failures observed during method development. They form a compact, extensible repair library rather than an exhaustive catalog and are frozen before evaluation. Appendix~\ref{app:principle-memory-entries} lists these seed entries and analyzes representative clean repair contrasts.

\subsection{Revision Operator}

The Revision Operator defines the local edit step in \textsc{SkillRevise}. 
Given the current skill $S_i$, Diagnosis $D_i$, and a set of bound principles $P_i$ supplied by the revision episode, it maps them to a proposed revised skill and a revision trace:
\begin{equation}
\label{eq:revision-operator}
(\hat S_{i+1}, z_i)=\mathcal{R}_\phi(S_i,D_i,P_i).
\end{equation}
The trace $z_i$ documents the evidence-to-edit mapping behind each revision, including the applied principles, targeted spans in $S_i$, relevant verification and preservation constraints, expected acceptance signals, and execution anchors.

An \emph{execution anchor} specifies how a textual edit should change executor behavior: the action to perform, the artifact or evidence to inspect, and where the instruction should appear in $\mathcal{I}(S_i)$. 
For example, an anchored JSON repair may require the executor to reload and parse the generated file, check required keys, and run local validation.

\subsection{Bounded Revision Episode}
\label{sec:revision-process}

A \textsc{SkillRevise} episode operationalizes the interfaces above under a finite revision budget. 
The budget constrains how many repair attempts may be evaluated, but it does not require the final generated candidate to be returned.
Given a task $T$, an initial skill $S_0$, an executor policy $\pi_\theta$, a fixed Principle Memory $\mathcal{M}$, and a revision budget $B$, the episode generates candidate revisions and returns the best observed skill under measured utility. 
Unless otherwise stated, the standard setting uses $B=3$ revision rounds; the execution protocol, pseudocode, and fixed implementation settings appear in Appendices~\ref{app:algorithmic-details} and~\ref{app:implementation-details}.

\paragraph{Episode state.}
The episode maintains a current skill $S_i$, a best observed skill $S_{\mathrm{best}}$, and an observed candidate set $\mathcal{H}$. 
Initially, $S_i=S_0$, $S_{\mathrm{best}}=S_0$, and $\mathcal{H}=\{S_0\}$. 
At each round, \textsc{SkillRevise} executes the current skill, diagnoses the resulting evidence, generates diagnosis-gated revisions when warranted, re-executes the candidate, and updates $S_{\mathrm{best}}$ only when measured utility improves.

The episode separates the search base from the returned artifact. 
If budget remains after generating $\hat S_{i+1}$, the next round sets $S_{i+1}\leftarrow \hat S_{i+1}$ so later diagnoses inspect the edited artifact. 
The returned artifact is instead the utility-gated best observed skill, not necessarily the latest generated revision.

\paragraph{Round transition.}
At round $i$, execution of the current skill produces
\begin{equation}
\label{eq:execution-evidence}
e_i=(\tau_i,v_i,r_i,c_i)=\Phi(T,S_i,\pi_\theta),
\end{equation}
where $\tau_i$ is the trajectory, $v_i$ is verifier feedback or evaluation evidence, $r_i$ is the outcome score or pass/fail reward, and $c_i$ records costs such as tokens, tool calls, steps, and latency.

\textsc{SkillRevise} constructs Diagnosis $D_i$ from $e_i$ and uses it as a gate for revision. 
If $A_i$ does not support a skill-level defect, the episode abstains from editing the skill for that evidence. 
When revision is warranted, \textsc{SkillRevise} retrieves a top-$m$ candidate set from Principle Memory:
\begin{equation}
\label{eq:principle-retrieval}
q_i=Q(T,D_i),\qquad C_i=\operatorname{Retrieve}_m(q_i,\mathcal{M}).
\end{equation}
The retrieval query uses task metadata and Diagnosis, including task family, acceptance criteria, diagnosis labels, causal judgment, rewrite targets, and evidence snippets. 
Hybrid retrieval combines sparse matching, which captures explicit labels, verifier terms, and rewrite targets, with dense matching, which captures semantically similar repair situations. 
Their rankings are combined by weighted reciprocal-rank fusion:
\begin{equation}
\label{eq:rrf}
\operatorname{score}(p)=
\frac{w_s}{\kappa+\operatorname{rank}_s(p)}
+\frac{w_d}{\kappa+\operatorname{rank}_d(p)},
\end{equation}
where $m$ is the retrieval width, distinct from the revision budget $B$, and $\kappa$ is the fusion constant.

Retrieval produces candidate principles, not final repair instructions. 
\textsc{SkillRevise} binds retrieved principles to the current diagnosis by retaining only those whose evidence requirements are satisfied and whose transfer constraints are not violated:
\begin{equation}
\label{eq:principle-binding}
P_i=\operatorname{Bind}(C_i,D_i).
\end{equation}
If no retrieved principle is supported, $P_i$ may be empty: Diagnosis remains available to guide revision, or the Diagnosis Gate may abstain when the evidence does not support a skill-level repair. A missing specialized principle can therefore reduce repair quality without forcing an unrelated repair.
The episode then calls the Revision Operator in Eq.~\ref{eq:revision-operator} to obtain a candidate revision $\hat S_{i+1}$ and trace $z_i$. 
The candidate is re-executed on the same task and added to the observed candidate set before utility-gated selection is applied.

\paragraph{Utility-gated selection.}
Let $\widehat{\mathcal{S}}_{\leq B}$ denote all generated and evaluated revision candidates within budget $B$. 
The observed set is
\begin{equation}
\mathcal{H}_{\leq B}=\{S_0\}\cup\widehat{\mathcal{S}}_{\leq B},
\end{equation}
and the selected artifact is
\begin{equation}
S^*_{\leq B}=\operatorname*{arg\,max}_{S\in\mathcal{H}_{\leq B}} U(S,T).
\end{equation}
Thus, increasing the revision budget expands the evaluated candidate set but does not imply that the final generated revision is selected. The utility may combine measured outcome, success-conditioned efficiency, transfer, and interference cost:
\begin{equation}
\begin{aligned}
U(S,T) &= \alpha\,\Delta_{\mathrm{succ}}(S,T) \\
       &\quad + \beta\,g_{\mathrm{succ}}(S,T)\Delta_{\mathrm{eff}}(S,T) \\
       &\quad + \gamma\,\Delta_{\mathrm{trans}}(S,\mathcal{F}) \\
       &\quad - \lambda\,C_{\mathrm{intf}}(S,\mathcal{F}).
\end{aligned}
\end{equation}
Here, $\Delta_{\mathrm{succ}}$, $\Delta_{\mathrm{eff}}$, and $\Delta_{\mathrm{trans}}$ denote changes in task outcome or success, execution efficiency, and transfer performance; $C_{\mathrm{intf}}$ measures interference on other tasks; and $\alpha,\beta,\gamma,\lambda$ weight the corresponding terms. 
The gate $g_{\mathrm{succ}}(S,T)$ conditions efficiency credit on successful execution. 
For binary verifiers, we set $g_{\mathrm{succ}}(S,T)=\mathbf{1}[\operatorname{succ}(S,T)=1]$, so lower-cost candidates are rewarded only after satisfying the verifier. 
This prevents the utility from favoring revisions that appear efficient simply because they stop early or omit required work.

\paragraph{Optional memory absorption.}
The dashed arrow in Figure~\ref{fig:skillrevise-pipeline} denotes optional post-evaluation or deployment-time memory maintenance. 
When enabled, an LLM abstracts validated repair patterns from calibration episodes into candidate principles; accepted entries are written to the bank, updating this non-parametric memory without changing model weights. Absorption is conservative: candidate principles must be evidence-backed, skill-level, utility-improving, and transferable, and filters remove task-specific constants, output paths, hidden answers, and verifier-specific shortcuts. Reported benchmark results use banks fixed before evaluation, and no evaluation-task feedback is written back during evaluation.
Appendix~\ref{app:prompts} lists the prompt interfaces.

\section{Experiments}

\begin{table*}[t]
\centering
\small
\setlength{\tabcolsep}{3.6pt}
{\renewcommand{\arraystretch}{0.94}
\resizebox{0.98\textwidth}{!}{%
\begin{tabular}{@{}llrrrrrrr@{}}
\toprule
\multirow{2}{*}{\textbf{Model}} & \multirow{2}{*}{\textbf{Method}} &
\multicolumn{2}{c}{\textbf{SkillsBench}} &
\multicolumn{2}{c}{\textbf{SkillLearn-R}} &
\multicolumn{2}{c}{\textbf{SWE-Skills-H}} &
\multirow{2}{*}{\textbf{Overall}} \\
\cmidrule(lr){3-4}\cmidrule(lr){5-6}\cmidrule(lr){7-8}
& & \textbf{Succ.} & \textbf{$\Delta$pp} & \textbf{Succ.} & \textbf{$\Delta$pp} & \textbf{Succ.} & \textbf{$\Delta$pp} & \\
\midrule
\multirow{5}{*}{\textbf{GPT-5.5}} & No skill & \succcell{31/86} & \basedelta & \succcell{20/50} & \basedelta & \succcell{28/70} & \basedelta & \overallcell{79/206} \\
& Skill-Creator & \succcell{34/86} & \deltapp{\gainpp{3.49}} & \succcell{17/50} & \deltapp{\losspp{-6.00}} & \succcell{27/70} & \deltapp{\losspp{-1.43}} & \overallcell{78/206} \\
& Skill $v_0$ & \succcell{35/86} & \deltapp{\gainpp{4.65}} & \succcell{23/50} & \deltapp{\gainpp{6.00}} & \succcell{29/70} & \deltapp{\gainpp{1.43}} & \overallcell{87/206} \\
& Rev. $v_1$ & \succcell{47/86} & \deltapp{\gainpp{18.60}} & \succcell{29/50} & \deltapp{\gainpp{18.00}} & \succcell{30/70} & \deltapp{\gainpp{2.86}} & \overallcell{106/206} \\
& \revsh{\textbf{Rev. $v_3$}} & \revsh{\bestsucc{53/86}} & \revsh{\deltapp{\gainpp{25.58}}} & \revsh{\bestsucc{29/50}} & \revsh{\deltapp{\gainpp{18.00}}} & \revsh{\bestsucc{33/70}} & \revsh{\deltapp{\gainpp{7.14}}} & \revsh{\bestoverall{115/206}} \\
\midrule
\multirow{5}{*}{\textbf{Opus-4.7}} & No skill & \succcell{16/86} & \basedelta & \succcell{7/50} & \basedelta & \succcell{32/70} & \basedelta & \overallcell{55/206} \\
& Skill-Creator & \succcell{28/86} & \deltapp{\gainpp{13.95}} & \succcell{7/50} & \deltapp{\zeropp} & \succcell{25/70} & \deltapp{\losspp{-10.00}} & \overallcell{60/206} \\
& Skill $v_0$ & \succcell{17/86} & \deltapp{\gainpp{1.16}} & \succcell{7/50} & \deltapp{\zeropp} & \succcell{23/70} & \deltapp{\losspp{-12.86}} & \overallcell{47/206} \\
& Rev. $v_1$ & \succcell{32/86} & \deltapp{\gainpp{18.60}} & \succcell{20/50} & \deltapp{\gainpp{26.00}} & \succcell{29/70} & \deltapp{\losspp{-4.29}} & \overallcell{81/206} \\
& \revsh{\textbf{Rev. $v_3$}} & \revsh{\bestsucc{44/86}} & \revsh{\deltapp{\gainpp{32.56}}} & \revsh{\bestsucc{25/50}} & \revsh{\deltapp{\gainpp{36.00}}} & \revsh{\succcell{31/70}} & \revsh{\deltapp{\losspp{-1.43}}} & \revsh{\bestoverall{100/206}} \\
\midrule
\multirow{5}{*}{\textbf{Kimi-2.6}} & No skill & \succcell{16/86} & \basedelta & \succcell{6/50} & \basedelta & \succcell{25/70} & \basedelta & \overallcell{47/206} \\
& Skill-Creator & \succcell{17/86} & \deltapp{\gainpp{1.16}} & \succcell{10/50} & \deltapp{\gainpp{8.00}} & \succcell{20/70} & \deltapp{\losspp{-7.14}} & \overallcell{47/206} \\
& Skill $v_0$ & \succcell{19/86} & \deltapp{\gainpp{3.49}} & \succcell{11/50} & \deltapp{\gainpp{10.00}} & \succcell{23/70} & \deltapp{\losspp{-2.86}} & \overallcell{53/206} \\
& Rev. $v_1$ & \succcell{28/86} & \deltapp{\gainpp{13.95}} & \succcell{15/50} & \deltapp{\gainpp{18.00}} & \succcell{25/70} & \deltapp{\zeropp} & \overallcell{68/206} \\
& \revsh{\textbf{Rev. $v_3$}} & \revsh{\bestsucc{37/86}} & \revsh{\deltapp{\gainpp{24.42}}} & \revsh{\bestsucc{22/50}} & \revsh{\deltapp{\gainpp{32.00}}} & \revsh{\bestsucc{27/70}} & \revsh{\deltapp{\gainpp{2.86}}} & \revsh{\bestoverall{86/206}} \\
\midrule
\multirow{5}{*}{\textbf{Qwen-3.6-Plus}} & No skill & \succcell{6/86} & \basedelta & \succcell{5/50} & \basedelta & \succcell{22/70} & \basedelta & \overallcell{33/206} \\
& Skill-Creator & \succcell{8/86} & \deltapp{\gainpp{2.33}} & \succcell{8/50} & \deltapp{\gainpp{6.00}} & \succcell{24/70} & \deltapp{\gainpp{2.86}} & \overallcell{40/206} \\
& Skill $v_0$ & \succcell{9/86} & \deltapp{\gainpp{3.49}} & \succcell{7/50} & \deltapp{\gainpp{4.00}} & \succcell{26/70} & \deltapp{\gainpp{5.71}} & \overallcell{42/206} \\
& Rev. $v_1$ & \succcell{24/86} & \deltapp{\gainpp{20.93}} & \succcell{12/50} & \deltapp{\gainpp{14.00}} & \succcell{34/70} & \deltapp{\gainpp{17.14}} & \overallcell{70/206} \\
& \revsh{\textbf{Rev. $v_3$}} & \revsh{\bestsucc{27/86}} & \revsh{\deltapp{\gainpp{24.42}}} & \revsh{\bestsucc{15/50}} & \revsh{\deltapp{\gainpp{20.00}}} & \revsh{\bestsucc{35/70}} & \revsh{\deltapp{\gainpp{18.57}}} & \revsh{\bestoverall{77/206}} \\
\midrule
\multirow{5}{*}{\textbf{DeepSeek-V4-Pro}} & No skill & \succcell{14/86} & \basedelta & \succcell{12/50} & \basedelta & \succcell{23/70} & \basedelta & \overallcell{49/206} \\
& Skill-Creator & \succcell{21/86} & \deltapp{\gainpp{8.14}} & \succcell{16/50} & \deltapp{\gainpp{8.00}} & \succcell{25/70} & \deltapp{\gainpp{2.86}} & \overallcell{62/206} \\
& Skill $v_0$ & \succcell{23/86} & \deltapp{\gainpp{10.47}} & \succcell{14/50} & \deltapp{\gainpp{4.00}} & \succcell{28/70} & \deltapp{\gainpp{7.14}} & \overallcell{65/206} \\
& Rev. $v_1$ & \succcell{34/86} & \deltapp{\gainpp{23.26}} & \succcell{23/50} & \deltapp{\gainpp{22.00}} & \succcell{29/70} & \deltapp{\gainpp{8.57}} & \overallcell{86/206} \\
& \revsh{\textbf{Rev. $v_3$}} & \revsh{\bestsucc{41/86}} & \revsh{\deltapp{\gainpp{31.40}}} & \revsh{\bestsucc{24/50}} & \revsh{\deltapp{\gainpp{24.00}}} & \revsh{\bestsucc{30/70}} & \revsh{\deltapp{\gainpp{10.00}}} & \revsh{\bestoverall{95/206}} \\
\bottomrule
\end{tabular}}}
\caption{Main results. Each benchmark cell reports success counts and absolute percentage-point gains ($\Delta$pp) over no-skill execution; Overall sums successes across the three benchmarks (206 tasks). SkillLearn-R denotes SkillLearnBench-Random; SWE-Skills-H denotes SWE-Skills-Bench-Hard. GPT-5.5 uses the Codex harness; other executors use Claude Code. Shading marks Rev. $v_3$; bold marks Rev. $v_3$ beating both baselines.}
\vspace{0.08in}
\label{tab:main-results}
\end{table*}

\subsection{Experimental Setup}
\label{sec:experimental_setup}

We evaluate \textsc{SkillRevise} as an execution-grounded skill revision framework on three verifier-driven benchmarks: the original SkillsBench, a 50-task SkillLearnBench-Random suite, and a 70-task SWE-Skills-Bench-Hard suite. Each benchmark uses a revision budget of three. For these three main benchmarks, unless otherwise stated, skill authoring and revision use GPT-5.5, while task execution is evaluated across GPT-5.5~\citep{openai2026gpt55}, Claude Opus 4.7~\citep{anthropic2026opus47}, Kimi K2.6~\citep{moonshotai2026kimik26}, Qwen3.6-Plus~\citep{alibabacloud2026qwen36plus}, and DeepSeek-V4-Pro~\citep{deepseekai2026deepseekv4pro}. We additionally report calibration-then-freeze principle-absorption studies on ALFWorld~\citep{shridhar2021alfworld} and BFCL-v4~\citep{patil2025bfcl} in Section~\ref{sec:domain-absorption}.

\setupheading{Evaluation conditions.}
We compare five evaluation conditions under the same task interface, executor, workspace, and verifier, so differences come from the installed skill artifact and revision budget. This controlled comparison isolates the effect of the installed skill artifact: no guidance, one-shot skill authoring, or execution-grounded revision. \textit{No skill} runs the executor without any skill artifact. \textit{Skill-Creator} installs a one-shot LLM-authored skill but performs no execution-grounded revision. \textit{Skill} $v_0$ is the initial skill produced by the \textsc{SkillRevise} authoring stage before observing verifier feedback, diagnosis, or revision. These first three conditions define the pre-revision baselines. \textit{Revision v1} and \textit{Revision v3} denote budgeted revision conditions with one and three allowed repair rounds, respectively. Reported runs use terminal-on-success stopping for cost control, so no further candidates are generated after the first verifier pass; final selection remains utility-gated over the resulting evaluated history. Thus, v1/v3 are budget-selected outcomes, not necessarily the most recently generated revision.

\setupheading{Benchmark protocol.}
We repackage SWE-Skills-Bench-Hard~\citep{han2026sweskillsbench} and SkillLearnBench-Random~\citep{zhong2026skilllearnbench} into the same task-directory interface used by SkillsBench. For both adapted benchmarks, we preserve the original task instructions, environments, and test-based evaluators where applicable. Task success is the primary metric. We also report outcome score, utility, tokens, tool calls, steps, and latency. Unless explicitly stated, reported benchmark results use a fixed Principle Memory available before evaluation, and do not use online absorption from test tasks. 
Appendices~\ref{app:benchmark-adaptation} and~\ref{app:manual-leakage-audit} give task-bundle records and the retained-skill leakage boundary.

\setupheading{Domain-specific absorption setup.}
For ALFWorld, we use 25 calibration tasks to absorb a compact 10-principle bank, then evaluate on 100 valid-seen and valid-unseen tasks. Qwen3-8B~\citep{qwen2025qwen3} is the executor. The initial skill $v_0$ is evaluated without the bank; later revisions retrieve from it. Evaluation tasks remain held out from online absorption.
For BFCL-v4, we use Qwen3-8B and the same calibration-then-freeze protocol separately for single-turn (ST) and multi-turn-base (MT-base) function calling. The 6- and 5-principle banks are frozen for their 100-task evaluations.

\subsection{Main Results}
\textbf{\textsc{SkillRevise} delivers broad gains over both no-skill execution and one-shot skill generation.} Table~\ref{tab:main-results} compares no-skill execution, one-shot Skill-Creator skills, the initial $v_0$ skill, and selected revised skills after one and three revision rounds. Across the 206 evaluated tasks, Revision v3 consistently improves overall success relative to no-skill execution: from 79 to 115 for GPT-5.5, 55 to 100 for Opus-4.7, 47 to 86 for Kimi-2.6, 33 to 77 for Qwen-3.6-Plus, and 49 to 95 for DeepSeek-V4-Pro. For the top executor, GPT-5.5, the gain spans all three benchmarks: Revision v3 raises SkillsBench success from 31/86 to 53/86, SkillLearnBench-Random from 20/50 to 29/50, and SWE-Skills-Bench-Hard from 28/70 to 33/70. The effect is not confined to GPT-5.5: Kimi-2.6, Qwen-3.6-Plus, and DeepSeek-V4-Pro improve over no skill on all three benchmarks, while Opus-4.7 achieves large gains on SkillsBench and SkillLearnBench-Random and remains close to no-skill performance on SWE-Skills-Bench-Hard.

\textbf{One-shot skill generation is not a reliable substitute for revision.} Skill-Creator sometimes helps, but its effect is inconsistent: it is slightly worse than no skill overall for GPT-5.5 (78/206 versus 79/206), tied overall for Kimi-2.6, and below no skill on several benchmark cells, including GPT-5.5 SkillLearnBench-Random and SWE-Skills-Bench-Hard. The initial $v_0$ skill is also not enough by itself; for example, Opus-4.7 drops from 55/206 without skills to 47/206 with $v_0$. In contrast, Revision v3 is the strongest overall condition for every executor in Table~\ref{tab:main-results}. The matched $v_0$-to-Revision comparisons therefore isolate the aggregate benefit of the complete execution-grounded revision lifecycle beyond initial skill conditioning; the component ablations below examine the contribution of its individual mechanisms.

\textbf{Most recovery appears early, while later rounds expand the candidate pool.} Revision v1 already improves over $v_0$ in every model--benchmark setting in Table~\ref{tab:main-results}. On SkillsBench, GPT-5.5 rises from 40.7\% to 54.7\%, Opus-4.7 from 19.8\% to 37.2\%, Kimi-2.6 from 22.1\% to 32.6\%, Qwen-3.6-Plus from 10.5\% to 27.9\%, and DeepSeek-V4-Pro from 26.7\% to 39.5\%. Later rounds still matter because unresolved tasks receive additional verifier-tested candidates; for example, Opus-4.7 rises from 37.2\% to 51.2\% on SkillsBench and from 40.0\% to 50.0\% on SkillLearnBench-Random. The smaller gains on SWE-Skills-Bench-Hard suggest that repository-specific, long-horizon failures are harder to repair through generic skill revision. On GPT-5.5 SkillsBench, the selected artifact reaches 53/86 while using the same 36.6k deployment tokens as direct no-skill execution, and maxrev1 already reaches 47/86. Under the same 86-task GPT-5.5/Codex evaluation and aligned maxrev3 budget, our reproduced SkillHone~\citep{li2026skillhone} and SkillOpt~\citep{yang2026skillopt} baselines reach 42/86 and 46/86, respectively, compared with 53/86 for \textsc{SkillRevise}; \textsc{SkillRevise} also uses fewer repair tokens, tool calls, and execution time. Appendix~\ref{app:domain-breakdown} reports a domain-level breakdown, while Tables~\ref{tab:skillsbench-deployment-cost} and~\ref{tab:skillsbench-repair-cost} separate deployment from repair-search costs.

\section{Analysis and Discussion}

\begin{figure}[t]
\centering
\includegraphics[width=\columnwidth]{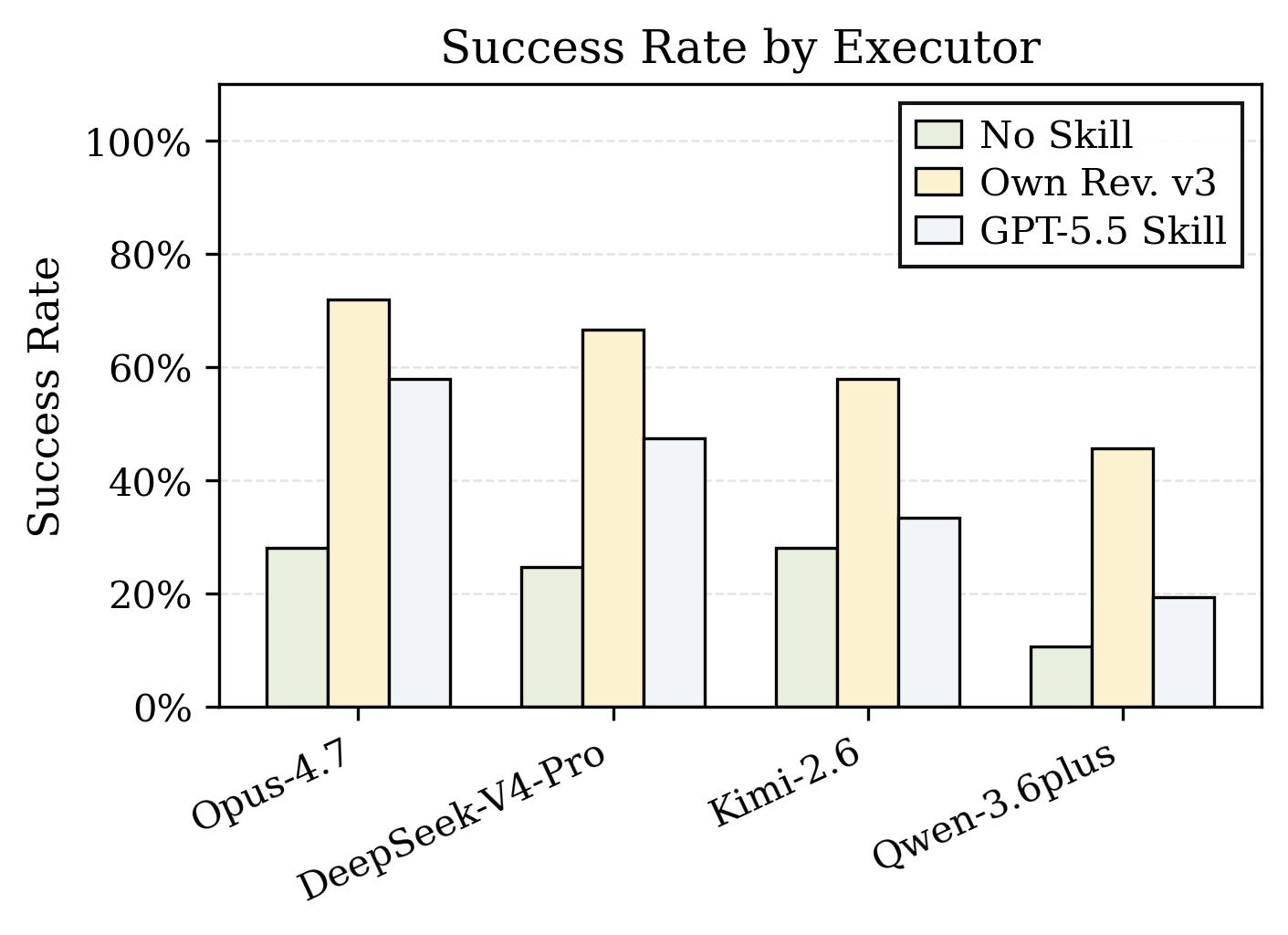}
\vspace{-0.3in}
\caption{Cross-model transfer on the 57-task GPT-5.5 source-success subset. Each group reports no-skill execution, executor-specific Revision v3, and fixed GPT-5.5-produced skills.}
\vspace{-0.1in}
\label{fig:transfer-app}
\end{figure}

\subsection{Analysis of Skills Transferability}

This analysis aims to examine whether revised skills encode transferable procedural knowledge, rather than serving as cue-based output of the executor model during revision. On the 57-task GPT-5.5 source-success subset, we compare three conditions for the four target executors with verified transfer runs: no-skill execution, the executor's own Revision v3 skill, and the same fixed GPT-5.5-produced skill artifact. Figure~\ref{fig:transfer-app} shows the four-executor comparison, and Appendix~\ref{app:transfer-details} reports the corresponding success counts.

The transfer results show partial but meaningful cross-executor portability. Across all four target executors, the fixed GPT-5.5-produced skill improves over no-skill execution, indicating that revision can produce procedural constraints and verifier-facing checks that are not tied exclusively to the source model. However, the magnitude of the transfer gain varies substantially, and executor-specific Revision $v_3$ is stronger in every case. This pattern suggests that revised skills include both reusable task execution paths and prompt conventions tailored to specific execution agents. The former reveals the generalizability of skill generation, while the latter motivates executor-aware revision.

\begin{table}[t]
\centering
\small
\begin{tabular*}{\linewidth}{@{\extracolsep{\fill}}lcccccc@{}}
\toprule
\textbf{Budget} & $v_0$ & $v_1$ & $v_2$ & $v_3$ & $v_4$ & $v_5$ \\
\midrule
maxrev1 & 55 & 31 & \missingcell & \missingcell & \missingcell & \missingcell \\
maxrev2 & 46 & 23 & 17 & \missingcell & \missingcell & \missingcell \\
maxrev3 & 42 & 19 & 13 & 12 & \missingcell & \missingcell \\
maxrev4 & 38 & 20 & 12 & 12 & 4 & \missingcell \\
maxrev5 & 37 & 20 & 11 & 11 & 4 & 3 \\
\bottomrule
\end{tabular*}
\caption{Selected skill versions by revision budget.}
\label{tab:selected}

\vspace{0.08in}

\centering
\footnotesize
\setlength{\tabcolsep}{3pt}
\begin{tabular*}{\linewidth}{@{\extracolsep{\fill}}lcc@{}}
\toprule
\textbf{Method} & \textbf{Succ.} & \textbf{Impact} \\
\midrule
No ablation & 53/86 & 0.00\% \\
w/o Diagnosis & 28/86 & $\downarrow$ 29.07\% \\
w/o Preserve Ledger & 42/86 & $\downarrow$ 12.79\% \\
w/o Execution Anchors & 44/86 & $\downarrow$ 10.47\% \\
w/o Principles & 45/86 & $\downarrow$ 9.30\% \\
Free-form Revision & 52/86 & $\downarrow$ 1.16\% \\
\midrule
Unstructured Repeated Revision ($v_3$) & 33/86 & $\downarrow$ 23.26\% \\
\bottomrule
\end{tabular*}
\caption{SkillsBench ablations and an unstructured control under a matched maxrev3 budget. Impact denotes the absolute success-rate drop from full \textsc{SkillRevise}.}
\label{tab:ablation}
\vspace{-0.1in}
\end{table}

\begin{table}[t]
\centering
\scriptsize
\setlength{\tabcolsep}{1.5pt}
\begin{tabular*}{\columnwidth}{@{\extracolsep{\fill}}lrrrrrr@{}}
\toprule
\textbf{Benchmark} & \textbf{\#P} & \textbf{No skill} & \textbf{$v_0$} & \textbf{Rev.} & \textbf{$\Delta$No} & \textbf{$\Delta v_0$} \\
\midrule
ALFWorld & 10 & 43/100 & 40/100 & \textbf{71/100} & +28 pp & +31 pp \\
BFCL-v4 ST & 6 & 82/100 & 83/100 & \textbf{91/100} & +9 pp & +8 pp \\
BFCL-v4 MT-base & 5 & 42/100 & 23/100 & \textbf{59/100} & +17 pp & +36 pp \\
\bottomrule
\end{tabular*}
\caption{Calibration-then-freeze results. \#P is the frozen-bank size; Rev. denotes \textsc{SkillRevise} maxrev3. ST/MT-base denote single-/multi-turn base.}
\vspace{-0.1in}
\label{tab:domain-absorption}
\end{table}

\subsection{Revision Rounds and Selection}
\label{sec:revision-rounds}

Revision rounds affect \textsc{SkillRevise} through both candidate generation and final selection. Here we focus on the selection behavior, while Appendix~\ref{app:revision-budget-sweep} reports the full success-rate and score sweep across revision caps. The budget maxrev$k$ allows up to $k$ attempted repairs, while the retained skill is chosen by the utility gate over the evaluated history. Reported runs stop generating candidates after a verifier pass. Increasing the budget therefore does not mean that the newest rewrite should automatically be deployed; it only expands the set of candidates that may be evaluated before stopping.

Table~\ref{tab:selected} shows this effect directly. With maxrev1, the selected artifact must come from either the initial $v_0$ skill or the first revision $v_1$. As the budget increases, later versions become available for tasks that remain unresolved, but early skills continue to dominate the retained set: even under maxrev5, the selector keeps $v_0$ for 37 tasks and $v_1$ for 20 tasks, while $v_4$ and $v_5$ account for only a small tail. This pattern supports the intended role of revision rounds as bounded search: later rounds add opportunities, terminal success can stop further search, and the utility gate prevents adoption based only on generation order.

\subsection{Component Ablations}

To separate framework structure from additional rewriting opportunities, we evaluate six within-framework variants and an additional unstructured repeated-revision baseline under the same $S_0$, executor, verifier, and maxrev3 budget. Free-form Revision retains the outer \textsc{SkillRevise} lifecycle but replaces its structured, principle-guided Revision Operator with free-form rewriting; its 52/86 result therefore isolates the operator's contribution.

Table~\ref{tab:ablation} shows that Diagnosis is the dominant component: removing it drops success from 53/86 to 28/86, indicating that revision needs an explicit bridge from evidence to skill-level defects. The preserve ledger, execution anchors, and Principle Memory yield smaller but substantial drops, while Free-form Revision remains close in raw success; the structured operator primarily improves grounding, attribution, and inspectability.

Unstructured Repeated Revision directly feeds leakage-safe verifier-visible feedback to the LLM to freely rewrite and re-execute the skill for three rounds, without the other \textsc{SkillRevise} components. Starting from the same $v_0$ (35/86), it achieves 33/86 after the third round, compared with 53/86 for full \textsc{SkillRevise} under maxrev3. Thus, additional feedback-driven rewriting and retry opportunities alone do not explain the improvement.

\subsection{Domain-Specific Principle Absorption}
\label{sec:domain-absorption}

To test Principle Memory beyond verifier-driven file tasks, we run ALFWorld with Qwen3-8B. Agents issue text commands rather than edit files or repositories, so success depends on admissible action syntax and state tracking. From 25 calibration tasks, we absorb a compact 10-principle bank covering recurring failures such as premature stopping, invalid actions, budget loss, and two-object goals, then evaluate on 100 valid-seen/unseen tasks.

The selected run solves 71/100 tasks, with 33/44 valid-seen and 38/56 valid-unseen successes. Table~\ref{tab:alfworld-family} in Appendix~\ref{app:alfworld-results} breaks this down by task family: success is highest on heat/place and light-inspection tasks, while clean/place remains hardest. The selector keeps $v_0$, $v_1$, and $v_2$ for 62, 31, and 7 tasks, respectively. Thus, the absorbed bank acts as targeted repair guidance rather than a default trigger, with principles used selectively rather than forcing revision on every episode.

BFCL-v4 tests whether the same lifecycle extends to function calling. With the frozen banks, \textsc{SkillRevise} reaches 91/100 on single-turn and 59/100 on multi-turn base, compared with 83/100 and 23/100 for $v_0$ and 82/100 and 42/100 without a skill. As Table~\ref{tab:domain-absorption} shows, it gains approximately 10 percentage points in the already strong single-turn setting. In ALFWorld and BFCL-v4 multi-turn base, where $v_0$ underperforms no-skill execution, revision reverses this negative transfer and improves over $v_0$ by more than 30 points.

\FloatBarrier

\section{Conclusion}

\textsc{SkillRevise} treats skills as execution-grounded artifacts that can be diagnosed, revised, and selected with verifier feedback. Across SkillsBench, SkillLearnBench-Random, and SWE-Skills-Bench-Hard, this bounded revision loop improves over both no-skill execution and one-shot Skill-Creator in most evaluated settings. The ALFWorld and BFCL-v4 studies further show that compact domain-specific principles acquired during calibration remain useful after the bank is frozen. The results suggest that reusable agent skills are most effective when they are not static advice, but testable procedural memory refined by observed failures.

\section*{Limitations}

\textsc{SkillRevise} depends on verifier-visible feedback and spends additional execution budget to obtain it. Sparse, flaky, or misaligned tests can lead revision to overfit visible checks, and some tasks remain better solved by direct no-skill exploration because maxrev3 does not fall back to no-skill execution. Our evaluation does not cover long-running or safety-critical deployments. Appendix~\ref{app:additional-limitations} discusses these limitations in more detail.

\section*{Acknowledgments}

The authors of this paper were supported by the National Key Research and Development Program of China (2025YFE0200500), the Beijing-Tianjin-Hebei Collaborative Innovation Special Project (26240201D), the ITSP Platform Research Project (ITS/189/23FP) from ITC of Hong Kong, SAR, China, and the AoE (AoE/E-601/24-N), the CRF (No. C6004-25G),  the RIF (R6021-20) and the GRF (16205322) from RGC of Hong Kong, SAR, China. The work described in this paper was conducted in full or in part by Dr. Haoran Li, JC STEM Early Career Research Fellow, supported by The Hong Kong Jockey Club Charities Trust.

\bibliography{custom}

\clearpage
\appendix
\raggedbottom
\setlength{\textfloatsep}{8pt plus 2pt minus 2pt}
\setlength{\floatsep}{6pt plus 2pt minus 2pt}
\setlength{\intextsep}{6pt plus 2pt minus 2pt}

\section{Benchmark Selection and Adaptation}
\label{app:benchmark-adaptation}

\subsection{Shared Bundle Interface}
SkillsBench is already close to the file-based task interface used by our harness, while SkillLearnBench-Random and SWE-Skills-Bench-Hard require adaptation from their native formats. We therefore convert all three sources into the same SkillsBench-style outer interface before running any skill condition. Each selected task directory contains \texttt{instruction.md}, \texttt{task.toml}, a verifier entrypoint under \texttt{tests/}, and task assets under \texttt{environment/} when needed. The released manifests record the final 86 SkillsBench, 50 SkillLearnBench-Random, and 70 SWE-Skills-Bench-Hard tasks. For the two adapted benchmarks, we first export benchmark-native tasks into this shared layout and then apply an explicit task-selection file; the runner reads the resulting manifest rather than relying on a hard-coded subset.

\subsection{SkillsBench}
The SkillsBench bundle contains the 86 selected tasks used for the main SkillsBench rows. We keep the task-facing instruction, environment files, and verifier contract in the shared bundle format, so the same runner can install no skill, a one-shot Skill-Creator artifact, or a revised \textsc{SkillRevise} artifact before invoking the verifier.

\subsection{SkillLearnBench-Random}
SkillLearnBench-Random~\citep{zhong2026skilllearnbench} is originally organized around task families with multiple instances and benchmark-specific evaluation scripts. We use a representative 50-task random subset that spans the main SkillLearnBench task families rather than concentrating on a single task type. We flatten the selected instances into independent task directories by copying each instance's instruction, metadata, environment assets, and tests into the shared bundle layout. The final subset is materialized from an explicit selection file, producing the manifest used for evaluation. We disable sibling-instance transfer context so that each selected instance is evaluated as a standalone task, and we score only verifier task success. The original SkillLearnBench coverage, safety, and executability LLM-judge metrics are not used in these comparisons, and oracle solution directories are not part of the public task bundle used for revision.

\subsection{SWE-Skills-Bench-Hard}
SWE-Skills-Bench-Hard~\citep{han2026sweskillsbench} is repository-based. Following the hard-setting motivation of SWE-Skills-Bench, we focus on tasks where direct LLM completion and self-generated skill use are weak, so the subset tests whether execution-grounded revision can improve skills in cases where simply adding a generated skill is not already sufficient. The final 70-task hard subset covers seven software-engineering skill families. The adaptation preserves the original software-engineering prompt, repository identity, configured environment, and task-specific tests while exposing them through the same task-directory interface. The exporter reads the benchmark configuration and task prompts, creates \texttt{instruction.md} and \texttt{task.toml}, builds the task environment from the original base image and repository commit, and copies the corresponding test files into the task verifier directory. Reference skills from the source benchmark are not exposed as solution artifacts during \textsc{SkillRevise}; the evaluated agent sees the task prompt, workspace, and verifier feedback under the same benchmark conditions.

\subsection{ALFWorld}
ALFWorld is not converted into the file-based task bundle above. It is used as an interactive environment for the principle-absorption study in Section~\ref{sec:domain-absorption}: a separate calibration split produces a compact principle bank, and the 100-task evaluation set is then executed without online absorption from evaluation tasks.

\subsection{BFCL-v4}
BFCL-v4~\citep{patil2025bfcl} is evaluated with Qwen3-8B through its executable function-calling interface. Single-turn evaluation uses the BFCL-v4 possible-answer files and official AST checker; multi-turn base uses \texttt{bfcl-eval} version 2026.3.23~\citep{mao2026bfcleval}, as recorded in the retained manifest. Separate single-turn and multi-turn-base calibration stages produce compact 6- and 5-principle banks, respectively; each bank is frozen during its 100-task evaluation, and evaluation feedback is not absorbed back into memory.

\section{Execution Protocol and Algorithm}
\label{app:algorithmic-details}

\subsection{Shared Evaluation Protocol}
For SkillsBench, SkillLearnBench-Random, and SWE-Skills-Bench-Hard, \textit{no skill}, \textit{Skill-Creator}, and \textsc{SkillRevise} use the same task interface, executor, verifier, and result aggregation whenever a matched run is available. \textit{No skill} executes without an installed skill. \textit{Skill-Creator} installs a one-shot LLM-authored skill and performs no execution-grounded revision. \textsc{SkillRevise} begins with its own initially authored skill $v_0$, generated before any execution feedback, then uses a fixed Principle Memory and up to three revision rounds. Reported runs stop generating further candidates after the first verifier pass and retain the utility-gated best observed skill from the resulting evaluated history. Test-task traces are not absorbed back into Principle Memory for later test tasks.

\subsection{GPT-5.5 SkillsBench Execution Profile}
For the GPT-5.5 SkillsBench rows, skill authoring, skill revision, and task execution all use GPT-5.5. The executor is the Codex ACP agent under the Docker BenchFlow backend, and every candidate is scored by the same task verifier. \textit{No skill} runs one verifier-scored execution without an installed skill. \textit{Skill-Creator} makes one LLM-authored skill and runs one skill-conditioned execution. \textsc{SkillRevise} first evaluates its initially authored $v_0$ skill, then attempts up to $B=3$ execution-grounded revision rounds; each evaluated candidate is re-executed before it can be selected. Under the shared protocol above, a maxrev3 run may retain $v_0$, $v_1$, $v_2$, or $v_3$, and selection never substitutes the no-skill baseline. The final artifact records one adopted trace per task, with status, reward, return code, timeout flag, tokens, tool calls, steps, and latency. For the 86-task GPT-5.5 SkillsBench run, the selector retains $v_0$, $v_1$, $v_2$, and $v_3$ on 42, 19, 13, and 12 tasks, respectively; the selected \textsc{SkillRevise} traces contain no timeout.

\begin{table}[!htbp]
\centering
\scriptsize
\setlength{\tabcolsep}{2.8pt}
\resizebox{\columnwidth}{!}{%
\begin{tabular}{@{}lrrrr@{}}
\toprule
\textbf{Deployment configuration} & \textbf{Succ.} & \textbf{Exec. tok.} & \textbf{Tools} & \textbf{Exec. lat.} \\
\midrule
Direct no-skill execution & 31/86 & 36.6k & 21.0 & 492.4s \\
Skill-Creator & 34/86 & 37.9k & 20.9 & 329.7s \\
\textbf{\textsc{SkillRevise} selected skill} & \textbf{53/86} & \textbf{36.6k} & 21.7 & 335.0s \\
\bottomrule
\end{tabular}}
\caption{One-shot deployment cost of the resulting GPT-5.5 SkillsBench artifact over 86 tasks. Costs refer to one execution of the displayed configuration after skill selection; they exclude preceding authoring and repair search. All reported costs are per-task averages.}
\label{tab:skillsbench-deployment-cost}
\end{table}

\begin{table}[!htbp]
\centering
\scriptsize
\setlength{\tabcolsep}{2.2pt}
\resizebox{\columnwidth}{!}{%
\begin{tabular}{@{}lrrrrr@{}}
\toprule
\textbf{Repair configuration} & \textbf{Succ.} & \textbf{Avg. repairs} & \textbf{Exec. tok.} & \textbf{Tools} & \textbf{Exec. lat.} \\
\midrule
\textsc{SkillRevise} maxrev1 & 47/86 & 0.59 & 23.9k & 14.0 & 204.4s \\
\textsc{SkillRevise} maxrev2 & 51/86 & 1.03 & 41.4k & 24.3 & 360.9s \\
\textbf{\textsc{SkillRevise} maxrev3} & \textbf{53/86} & 1.42 & 56.9k & 33.6 & 484.5s \\
SkillHone maxrev3 & 42/86 & 1.77 & 69.8k & 36.0 & 746.7s \\
SkillOpt maxrev3 & 46/86 & 1.73 & 73.6k & 39.1 & 685.3s \\
\bottomrule
\end{tabular}}
\caption{Post-initial repair-search cost on the same 86 SkillsBench tasks under terminal-on-success accounting. Costs exclude initial skill construction, the common initial $v_0$ execution, and controller-side revision generation. Each executed repair corresponds to one revision-generation model call; Diagnosis adds no separate LLM call in this configuration. Executor tokens, tool calls, and latency are per-task averages.}
\label{tab:skillsbench-repair-cost}
\end{table}

\noindent\textbf{Controller-side latency.}
Revision generation adds 23.1, 40.6, and 55.5 seconds per task for maxrev1, maxrev2, and maxrev3, respectively, yielding total repair-stage latencies of 227.5, 401.5, and 540.0 seconds. At maxrev3, revision generation accounts for 10.3\% of the combined repair-stage latency.

\noindent\textbf{Amortized break-even.}
Relative to $v_0$, maxrev1, maxrev2, and maxrev3 add 12, 16, and 18 observed successes. Maxrev1 captures 66.7\% of the maxrev3 gain using 42.0\% of its repair-execution tokens and 42.1\% of its total repair-stage latency; maxrev2 captures 88.9\% of the gain using 72.8\% of the tokens and 74.4\% of the latency. Across 86 tasks, the three budgets require approximately 171k, 223k, and 272k repair-execution tokens and 27, 36, and 43 minutes per additional observed success. Maxrev1 already exceeds both three-round revision baselines at lower overhead, while maxrev3 achieves the highest success and remains cheaper than both SkillHone and SkillOpt. A universal economic break-even depends on the value of a successful execution and how often the frozen skill is reused: revision is justified when accumulated future gains outweigh its one-time repair cost.

\subsection{Revision Budget Sweep}
\label{app:revision-budget-sweep}

Table~\ref{tab:rounds} reports the GPT-5.5 SkillsBench budget sweep that complements Section~\ref{sec:revision-rounds}.

\begin{table}[H]
\centering
\small
\begin{tabular*}{\columnwidth}{@{\extracolsep{\fill}}lrrrr@{}}
\toprule
\textbf{Budget} & \textbf{Succ.} & \textbf{Rate} & \textbf{Out.} & \textbf{Overall} \\
\midrule
maxrev1 & 47/86 & 54.65 & 0.577 & 0.138 \\
maxrev2 & 51/86 & 59.30 & 0.617 & 0.183 \\
maxrev3 & 53/86 & 61.63 & 0.644 & 0.213 \\
maxrev4 & 55/86 & 63.95 & 0.658 & 0.235 \\
maxrev5 & 56/86 & 65.12 & 0.669 & 0.248 \\
\bottomrule
\end{tabular*}
\caption{Effect of revision budget on SkillsBench.}
\label{tab:rounds}
\end{table}

\subsection{Bounded Revision Algorithm}
Algorithm~\ref{alg:skillrevise-app} expands the bounded revision episode used in the main experiments. The standard setting uses $B=3$ revision rounds, a fixed Principle Memory available before evaluation, utility-gated best-observed selection, and terminal-on-success stopping for the reported benchmark runs. Online absorption from evaluation tasks is disabled. The three main benchmarks use the fixed seven-principle seed memory, whereas ALFWorld and BFCL-v4 use separately calibrated banks that are frozen before evaluation.

\begin{algorithm}[!htbp]
\small
\caption{Maxrev3 \textsc{SkillRevise} episode}
\label{alg:skillrevise-app}
\begin{algorithmic}[1]
\Require task $T$, initial skill $S_0$, executor $\pi_\theta$, verifier $\Phi$, fixed Principle Memory $\mathcal{M}$, budget $B=3$
\State evaluate $S_0$: $e_0 \gets \Phi(T,S_0,\pi_\theta)$
\State initialize $S_{\mathrm{cur}}\gets S_0$, $S_{\mathrm{best}}\gets S_0$, and $\mathcal{H}\gets\{(S_0,e_0)\}$
\For{$i=0$ to $B-1$}
  \If{$e_i$ is terminal under the stopping rule}
    \State \textbf{break}
  \EndIf
  \State construct Diagnosis $D_i$ from the current evidence $e_i$
  \If{$D_i$ does not support a skill-level defect}
    \State record an abstention trace and stop revising
    \State \textbf{break}
  \EndIf
  \State retrieve candidates $C_i \gets \operatorname{Retrieve}_m(Q(T,D_i),\mathcal{M})$
  \State bind principles $P_i \gets \operatorname{Bind}(C_i,D_i)$ using evidence requirements and transfer constraints
  \State generate $(\hat S_{i+1},z_i)\gets\mathcal{R}_\phi(S_{\mathrm{cur}},D_i,P_i)$
  \State re-execute the candidate: $\hat e_{i+1}\gets\Phi(T,\hat S_{i+1},\pi_\theta)$
  \State append $(\hat S_{i+1},\hat e_{i+1},z_i)$ to $\mathcal{H}$
  \If{$U(\hat S_{i+1},T)>U(S_{\mathrm{best}},T)$}
    \State $S_{\mathrm{best}}\gets\hat S_{i+1}$
  \EndIf
  \State $S_{\mathrm{cur}}\gets\hat S_{i+1}$ and $e_{i+1}\gets\hat e_{i+1}$
\EndFor
\State \Return $S_{\mathrm{best}}$ and the full candidate history $\mathcal{H}$
\end{algorithmic}
\end{algorithm}

The revision trace $z_i$ is stored with the candidate and contains the verifier contract, failure and preserve ledgers, selected and ignored principles, execution anchors, and acceptance signals.

\section{Implementation Settings}
\label{app:implementation-details}

This appendix records the fixed runtime and retrieval settings needed to reproduce the reported \textsc{SkillRevise} runs. Benchmark task bundles and selection records are described in Appendix~\ref{app:benchmark-adaptation}; prompt interfaces are listed in Appendix~\ref{app:prompts}.

\paragraph{Execution and selection.}
For the three main benchmarks, unless otherwise stated, skill authoring and revision use GPT-5.5, while the task executor is the model named in the corresponding result row. Each task starts from an initially authored skill $v_0$. Execution and selection follow Algorithm~\ref{alg:skillrevise-app}; budget sweeps vary only $B$.

\paragraph{Retrieval and binding.}
Principle Memory is fixed before evaluation for SkillsBench, SkillLearnBench-Random, and SWE-Skills-Bench-Hard; online absorption from test tasks is disabled. At each revision step, Diagnosis supplies task metadata, acceptance criteria, labels, causal judgment, rewrite targets, and evidence snippets to a hybrid retriever. The revision prompt receives a small candidate set and must select only evidence-supported principles.
ALFWorld and BFCL-v4 instead use calibration-built banks that remain frozen during evaluation. In the SkillsBench traces, 44 of the 50 tasks that generate revisions select at least one principle, all seven seed principles are used, and six tasks revise without selecting a principle; these counts measure utilization, not memory coverage.

\paragraph{Runtime logging.}
The LLM wrapper uses deterministic low-temperature generation and retry logic; task-level verifier timeouts are inherited from task metadata. Each evaluated candidate records status, reward, return code, timeout flag, token count, tool calls, steps, latency, selected principles, ignored principles, and execution anchors.

\paragraph{Fixed settings.}
The seed memory contains seven repair principles. Retrieval combines sparse and dense rankings by reciprocal-rank fusion with $w_s=0.5$, $w_d=0.5$, and $\kappa=60$. Dense scoring uses \texttt{qwen/qwen3-embedding-4b} with metadata-dominant retrieval and content weight 0.05. Up to four principles are injected into the revision prompt, and at most three may be selected. The wrapper uses a 600s HTTP timeout, three retries with 2s base delay, temperature 0.2, and 4096 maximum generation tokens. ALFWorld uses a 10-principle bank built from 25 calibration tasks; BFCL-v4 uses separate frozen banks with 6 principles for single-turn and 5 for multi-turn base.

\section{Cross-Model Transfer Details}
\label{app:transfer-details}

Table~\ref{tab:transfer-details} gives the success counts behind Figure~\ref{fig:transfer-app} for the four target executors with verified transfer runs. All results use the same 57-task GPT-5.5 source-success subset. \emph{Own Rev. v3} reports the executor-specific Revision v3 result on this subset. \emph{GPT-5.5 skill} installs the fixed GPT-5.5-produced final skill artifact for the executor.

\begin{center}
\small
\resizebox{\columnwidth}{!}{%
\begin{tabular}{@{}lccc@{}}
\toprule
\textbf{Executor} & \textbf{No skill} & \textbf{Own Rev. v3} & \textbf{GPT-5.5 skill} \\
\midrule
Opus-4.7 & 16/57 (28.07) & 41/57 (71.93) & 33/57 (57.89) \\
DeepSeek-V4-Pro & 14/57 (24.56) & 38/57 (66.67) & 27/57 (47.37) \\
Kimi-2.6 & 16/57 (28.07) & 33/57 (57.89) & 19/57 (33.33) \\
Qwen-3.6-Plus & 6/57 (10.53) & 26/57 (45.61) & 11/57 (19.30) \\
\bottomrule
\end{tabular}}
\end{center}
\refstepcounter{table}\label{tab:transfer-details}
\noindent\begin{minipage}{\columnwidth}
\footnotesize Table~\thetable: Data for Figure~\ref{fig:transfer-app}. Values are successes over the 57-task GPT-5.5 source-success subset; parentheses give rates.
\end{minipage}

\paragraph{Analysis.}
The fixed GPT-5.5-produced skills improve over no-skill execution for all four target executors, but the size of the transfer gain varies substantially. The largest gains appear for Opus-4.7 (+17 successes) and DeepSeek-V4-Pro (+13 successes), while Kimi-2.6 and Qwen-3.6-Plus show smaller gains of +3 and +5 successes. This indicates that revised skills contain reusable procedural guidance, but their utility depends on how well the target executor follows the source skill's tool-use and verification conventions. Executor-specific revision remains stronger in every row: Own Rev. v3 exceeds the transferred GPT-5.5 skill by 8 successes for Opus-4.7, 11 for DeepSeek-V4-Pro, 14 for Kimi-2.6, and 15 for Qwen-3.6-Plus. Thus, transfer signals portability, but executor-specific revision remains important.

\section{Domain-Level SkillsBench Breakdown}
\label{app:domain-breakdown}

Table~\ref{tab:domain-breakdown} compares SkillsBench domain-level success counts for GPT-5.5 and Opus-4.7 under no skill, one-shot Skill-Creator, and Revision v3.

\begin{center}
\scriptsize
\setlength{\tabcolsep}{2pt}
\resizebox{\columnwidth}{!}{%
\begin{tabular}{@{}lcccccc@{}}
\toprule
& \multicolumn{3}{c}{\textbf{GPT-5.5}} & \multicolumn{3}{c}{\textbf{Opus-4.7}} \\
\cmidrule(lr){2-4}\cmidrule(l){5-7}
\textbf{Domain} & \textbf{No} & \textbf{Creator} & \textbf{Rev} & \textbf{No} & \textbf{Creator} & \textbf{Rev} \\
\midrule
Energy & 1/4 & 1/4 & 3/4 & 1/4 & 1/4 & 3/4 \\
Office & 6/14 & 4/14 & 9/14 & 2/14 & 3/14 & 8/14 \\
Math & 5/10 & 6/10 & 8/10 & 4/10 & 5/10 & 7/10 \\
Robotics & 2/6 & 5/6 & 6/6 & 1/6 & 5/6 & 3/6 \\
Software & 3/11 & 5/11 & 8/11 & 1/11 & 4/11 & 5/11 \\
Science & 5/13 & 4/13 & 7/13 & 5/13 & 6/13 & 9/13 \\
Media & 5/11 & 3/11 & 4/11 & 0/11 & 1/11 & 4/11 \\
Finance & 2/7 & 3/7 & 4/7 & 1/7 & 1/7 & 2/7 \\
Cyber. & 2/6 & 3/6 & 3/6 & 1/6 & 2/6 & 3/6 \\
Health. & 0/1 & 0/1 & 0/1 & 0/1 & 0/1 & 0/1 \\
Manuf. & 0/3 & 0/3 & 1/3 & 0/3 & 0/3 & 0/3 \\
\bottomrule
\end{tabular}}
\end{center}
\refstepcounter{table}\label{tab:domain-breakdown}
\noindent\begin{minipage}{\columnwidth}
\footnotesize Table~\thetable: Domain-level SkillsBench success counts for GPT-5.5 and Opus-4.7. Each cell reports successful tasks over evaluated tasks. Creator denotes the Skill-Creator baseline, Rev denotes Revision v3, and Cyber., Health., and Manuf. abbreviate Cybersecurity, Healthcare, and Manufacturing.
\end{minipage}

\paragraph{Analysis.}
The domain breakdown shows that the aggregate SkillsBench gains are not concentrated in a single task family. For GPT-5.5, Revision v3 improves over no-skill execution in 9 of 11 domains, accounting for the overall gain from 31 to 53 successes. The largest absolute increases occur in Software (3/11 to 8/11) and Robotics (2/6 to 6/6), followed by Office and Math (+3 successes each). This suggests that revision is especially useful when failures come from missing procedural steps, fragile tool use, or incomplete verifier-facing checks. The main exception is Media, where GPT-5.5 drops from 5/11 without skills to 4/11 after revision, indicating that skill conditioning can still hurt when direct exploration is already competitive.

The Opus-4.7 breakdown shows a similar but not identical pattern. Revision v3 improves over no-skill execution in 9 of 11 domains and yields the largest gains in Office (2/14 to 8/14), Science (5/13 to 9/13), Media (0/11 to 4/11), and Software (1/11 to 5/11). These improvements explain why Opus-4.7 has the largest aggregate SkillsBench gain in Table~\ref{tab:main-results}. However, the Robotics row also shows a useful failure case: Skill-Creator reaches 5/6 while Revision v3 reaches 3/6, so iterative revision is not uniformly dominant at the domain level. This reinforces the need for explicit selection and for analyzing domain-level regressions rather than relying only on aggregate success.

\section{Case Studies}
\label{app:case-studies}

We analyze representative GPT-5.5 SkillsBench tasks because this setting has matched no-skill, Skill-Creator, and maxrev3 \textsc{SkillRevise} runs for all evaluated tasks. Let $B$ denote the no-skill baseline, $C$ denote Skill-Creator, and $R$ denote \textsc{SkillRevise} maxrev3. Table~\ref{tab:case-studies} summarizes three representative outcome patterns: recovery when both baselines fail, repair of a skill-induced failure, and residual skill-induced regression. Here, $R$ denotes the utility-selected skill within the revision budget; selection ranges only over skill versions and does not include a no-skill fallback.

\begin{table}[H]
\centering
\scriptsize
\setlength{\tabcolsep}{2pt}
\begin{tabular*}{\columnwidth}{@{\extracolsep{\fill}}llll@{}}
\toprule
\textbf{Pattern} & \textbf{Count} & \textbf{Representative task} & \textbf{Selected} \\
\midrule
$B=0,C=0,R=1$ & 13/86 & react-performance-debugging & $v_1$ \\
$B=1,C=0,R=1$ & 7/86 & exceltable-in-ppt & $v_1$ \\
$B=1,C=0,R=0$ & 2/86 & video-filler-word-remover & $v_0$ \\
\bottomrule
\end{tabular*}
\caption{Representative GPT-5.5 SkillsBench case-study patterns. $B$ is no skill, $C$ is Skill-Creator, and $R$ is \textsc{SkillRevise} maxrev3.}
\label{tab:case-studies}
\end{table}

\begin{table}[H]
\centering
\scriptsize
\setlength{\tabcolsep}{3pt}
\begin{tabular*}{\columnwidth}{@{}p{0.22\columnwidth}@{\extracolsep{\fill}}p{0.72\columnwidth}@{}}
\toprule
\textbf{Run} & \textbf{Evidence and revision signal} \\
\midrule
No skill & \textbf{Fail.} The agent does not satisfy the checkout-latency verifier, and no reusable non-regression checklist is available to prioritize the failing performance contract. \\
Skill-Creator & \textbf{Fail.} The one-shot skill gives broad Next.js optimization advice but remains too generic, with brittle literals and no bounded fallback path. \\
\textsc{SkillRevise} $v_0$ & \textbf{Fail.} Checkout takes 916ms, above the required 800ms threshold, while UI, bundle, rerender, cart, compare-page, and test-id checks pass. Diagnosis labels false certainty, over-specificity, context pollution, and wrong abstraction level while preserving the passed checks. \\
\textsc{SkillRevise} $v_1$ & \textbf{Pass; selected.} The verifier succeeds with outcome score 1.0. The revised skill adds endpoint-specific measurement, fallback timing checks, preserve constraints for test-observed UI behavior, and a smallest-failing-check rerun before broader verification. \\
\bottomrule
\end{tabular*}
\caption{Trace of the web-performance case study.}
\label{tab:react-trace}
\end{table}

\clearpage
\caseheading{Worked example: a revised skill succeeds.}
In \texttt{react-performance-debugging}, the initial skill $v_0$ fails the checkout-latency verifier: the checkout path takes 916ms, above the required 800ms threshold, even though the UI, bundle, rerender, cart, compare-page, and test-id checks pass. Table~\ref{tab:react-trace} shows the corresponding version-level trace. Diagnosis attributes the failure to a skill that is too literal and too certain, with no fallback path when the first optimization plan is insufficient. The selected revised skill $v_1$ succeeds with outcome score 1.0 and overall utility 1.042.

\begin{tcolorbox}[colback=promptback, colframe=promptframe, colbacktitle=promptframe, coltitle=white, fonttitle=\bfseries\small, title={Successful revised skill excerpt}]
\small
\textbf{Selected version:} $v_1$, \emph{Next.js Web Performance Non-Regression Workflow}.

\textbf{Core revised checkpoints:}
\begin{itemize}[leftmargin=*,nosep]
\item Discover project contracts before editing: scripts, tests, routes, components, timing thresholds, and invariants that already pass.
\item Localize the bottleneck with measured evidence from tests, request timing, build output, logs, or profiling; fall back to the closest available check when a tool is missing.
\item Apply the smallest targeted fix, such as removing artificial waits, repeated expensive work, sequential independent awaits, oversized payloads, or per-request recomputation.
\item Verify in increasing scope: rerun the smallest latency check first, then broader build and verifier tests, while preserving \mbox{\texttt{data-testid} attributes and instrumentation}.
\end{itemize}
\end{tcolorbox}

\caseheading{Grounded revision recovers failures.}
In \texttt{react-performance-debugging}, both no-skill execution and Skill-Creator fail, while \textsc{SkillRevise} succeeds with $v_1$. The failed initial traces expose a typical skill-design problem: the guidance is too generic for a Next.js performance task that must simultaneously improve latency, preserve behavior, and pass UI tests. Diagnosis flags missing fallback handling, over-specific literals, context pollution, and false certainty. The selected revision reframes the skill as a non-regression workflow: discover local scripts and tests, localize the bottleneck with measurable evidence, preserve test IDs and user-facing behavior, and rerun the relevant verifier before finalizing. This case illustrates the core benefit of revision: verifier feedback turns a broad optimization skill into an execution checklist that constrains the agent toward targeted, behavior-preserving edits.

\caseheading{Revision can repair skill-induced errors even when no skill already succeeds.}
In \texttt{exceltable-in-ppt}, no-skill execution succeeds, Skill-Creator fails, and \textsc{SkillRevise} succeeds with $v_1$. The Skill-Creator and initial $v_0$ failures are not due to lack of task ability; the verifier reports that an unrelated spreadsheet cell is corrupted, including a non-target CNY-to-EUR value changing from 0.127 to \texttt{NaN}. The revision therefore adds operational safeguards: snapshot embedded workbook cells before editing, identify the exact row and column from slide evidence, update only the target non-formula cell, compare the reopened output against the pre-write snapshot, and retry with a lower-impact workbook-part replacement if serialization corrupts unrelated cells. This pattern shows how \textsc{SkillRevise} reduces skill-induced risk by turning a high-level edit recipe into a preservation-aware procedure.

\caseheading{Skill guidance can still harm no-skill successes.}
In \texttt{video-filler-word-remover}, no-skill execution succeeds, but both Skill-Creator and \textsc{SkillRevise} fail. The verifier indicates that the produced filler-word clip duration falls outside the accepted range, so the failure is a calibration error in segment detection rather than a missing output file. Across the revision episode, the method repeatedly diagnoses context pollution and false certainty and attempts to add validation and fallback rules, but no revised candidate succeeds and the fallback remains $v_0$. This is a residual limitation of skill-conditioned execution: when a task can be solved by direct instance-specific exploration, a generic media-processing workflow can over-regularize the agent into a brittle ASR/timestamping pipeline. Because maxrev3 does not fall back to no-skill execution, this case remains a failure for \textsc{SkillRevise}.

\section{Domain-Specific Principle Absorption}
\label{app:alfworld-results}

Table~\ref{tab:alfworld-family} reports the task-family success rates for the ALFWorld run in Section~\ref{sec:domain-absorption}. The totals match the selected 71/100 run reported in the main text.

\begin{table}[H]
\centering
\scriptsize
\setlength{\tabcolsep}{3pt}
\begin{tabular*}{\columnwidth}{@{\extracolsep{\fill}}lrr@{}}
\toprule
\textbf{ALFWorld task family} & \textbf{Success} & \textbf{SR (\%)} \\
\midrule
Look at object in light & 18/21 & 85.7 \\
Pick and place & 13/19 & 68.4 \\
Clean then place & 6/15 & 40.0 \\
Cool then place & 13/18 & 72.2 \\
Heat then place & 14/16 & 87.5 \\
Pick two and place & 7/11 & 63.6 \\
\midrule
Overall & 71/100 & 71.0 \\
\bottomrule
\end{tabular*}
\caption{ALFWorld task-family success rates for \textsc{SkillRevise} with Qwen3-8B on 100 evaluation tasks.}
\label{tab:alfworld-family}
\end{table}

Table~\ref{tab:alfworld-comparison} compiles reported ALFWorld results across heterogeneous settings. The Reflexion, ReflAct, GiGPO, and constant-context SFT+RL results are taken from their respective papers. The GRPO result is reported by GiGPO, while the SimpleMem+GRPO, Mem0+GRPO, ExpeL, EvolveR, Memp, and Mem0 results are taken from Table~1 of SkillRL~\citep{xia2026skillrl}.

\begin{table}[H]
\centering
\scriptsize
\setlength{\tabcolsep}{2.2pt}
\resizebox{\columnwidth}{!}{%
\begin{tabular}{@{}lrl@{}}
\toprule
\textbf{Method} & \textbf{Reported SR} & \textbf{Setting} \\
\midrule
Reflexion~\citep{shinn2023reflexion} & 97.0 & GPT-3; 134 tasks; up to 12 trials \\
ReflAct~\citep{kim2025reflact} & 93.3 & GPT-4o; 134 test tasks; prompt-only \\
GiGPO (w/ std)~\citep{feng2025group} & 90.8$\pm$1.3 & Qwen2.5-7B; RL; 3 seeds \\
Constant-context SFT+RL~\citep{xie2026history} & 89.6$\pm$0.4 & Qwen3-8B; 134 unseen tasks \\
GRPO~\citep{feng2025group} & 77.6$\pm$5.2 & Qwen2.5-7B; RL; 3 seeds \\
\textbf{\textsc{SkillRevise}} & 71.0 & Ours; Qwen3-8B; 25 cal./100 eval. \\
SimpleMem+GRPO~\citep{liu2026simplemem} & 62.5 & Qwen2.5-7B \\
Mem0+GRPO~\citep{chhikara2025mem0} & 54.7 & Qwen2.5-7B \\
ExpeL~\citep{zhao2024expel} & 46.3 & Qwen2.5-7B \\
EvolveR~\citep{wu2025evolver} & 43.8 & Qwen2.5-7B \\
Memp~\citep{fang2026memp} & 41.4 & Qwen2.5-7B \\
Mem0~\citep{chhikara2025mem0} & 33.6 & Qwen2.5-7B \\
\bottomrule
\end{tabular}}
\caption{Reported ALFWorld results are not directly comparable because models, task sets, training budgets, and trial counts differ.}
\label{tab:alfworld-comparison}
\end{table}

\section{Failure Modes}
\label{app:failure-modes}

Against Skill-Creator, the standard \textsc{SkillRevise} maxrev3 setting wins by success on 20 tasks, ties on 65, and loses on 1; by overall utility, it wins on 58 tasks, ties on 2, and loses on 26. The gap between success and overall utility reflects an important failure mode: some revised skills solve the task but increase execution cost. Other failures arise when Diagnosis attributes environment or model-capability failures to the skill, when a revision overfits to the observed verifier, or when the required fix is a specialized domain method absent from the current Principle Memory.

\subsection{Quantified failure taxonomy}
We summarize the GPT-5.5 SkillsBench maxrev3 revision traces to characterize recurring failure types. Across 86 tasks, the trace set contains 208 candidate-level diagnosis entries. The coarse diagnosis labels are dominated by skill-form defects: \texttt{context\_pollution} appears in 207 entries, \texttt{false\_certainty} in 181, and both \texttt{over\_specificity} and \texttt{wrong\_abstraction\_level} in 54 entries; \texttt{environment\_mismatch} appears once. Table~\ref{tab:failure-taxonomy} reports the more operational failure types stored in the structured failure ledgers for the 122 in-budget revision operations that produced the evaluated \(v_1\)--\(v_3\) candidates. Primary counts are single-label, while secondary mentions are multi-label and therefore do not sum to the number of records.

\newpage
\subsection{Skill-induced regression}
Using Boolean task success, one-shot Skill-Creator regresses 9 of the 31 no-skill successes.

\begin{table}[H]
\centering
\scriptsize
\setlength{\tabcolsep}{2.5pt}
\renewcommand{\arraystretch}{1.08}
\begin{tabular*}{\columnwidth}{@{}>{\raggedright\arraybackslash}p{0.60\columnwidth}@{\extracolsep{\fill}}rrr@{}}
\toprule
\textbf{Failure type and typical signal} & \textbf{Prim.} & \textbf{Sec.} & \textbf{Tasks} \\
\midrule
{\textbf{Method / algorithmic workflow}\newline\emph{Wrong solver, media, numeric, or artifact-generation procedure.}} & 45 & 23 & 34 \\
\addlinespace[1pt]
{\textbf{Verification gap}\newline\emph{Output partly passes tests but misses a verifier-critical check.}} & 18 & 65 & 41 \\
\addlinespace[1pt]
{\textbf{Schema / artifact format}\newline\emph{Invalid JSON/CSV/XLSX structure, fields, ranges, or serialization.}} & 11 & 22 & 20 \\
\addlinespace[1pt]
{\textbf{Constraint / efficiency cost}\newline\emph{Preserved behavior, word budget, token cost, or latency is violated.}} & 11 & 5 & 9 \\
\addlinespace[1pt]
{\textbf{Data understanding}\newline\emph{Semantic answer, counting, extraction, or interpretation is wrong.}} & 8 & 8 & 9 \\
\addlinespace[1pt]
{\textbf{Overfit / hard-code}\newline\emph{Repair depends on over-specific literals or instance-shaped shortcuts.}} & 2 & 10 & 9 \\
\addlinespace[1pt]
{\textbf{Incomplete execution}\newline\emph{Required file, cell range, or processing stage is not produced.}} & 1 & 0 & 1 \\
\addlinespace[1pt]
{\textbf{Environment / tool / path}\newline\emph{Tool, environment, or output-path alignment is implicated.}} & 0 & 3 & 3 \\
\addlinespace[1pt]
{\textbf{Unknown / untyped}\newline\emph{Revision metadata lacks a structured failure ledger.}} & 26 & 0 & 17 \\
\bottomrule
\end{tabular*}
\caption{Failure taxonomy for GPT-5.5 SkillsBench maxrev3 revision attempts. Primary counts cover 122 in-budget revisions producing evaluated \(v_1\)--\(v_3\) candidates; secondary counts are multi-label; Tasks counts unique tasks with any mention.}
\label{tab:failure-taxonomy}
\end{table}
\FloatBarrier

\noindent\textsc{SkillRevise} maxrev3 reduces this to 2 no-skill successes that become failures, and it regresses 1 Skill-Creator success. These remaining regressions are the cases where a skill-conditioned execution is still worse than direct task exploration, motivating fallback-aware selection or routing in future work.

\clearpage
\twocolumn[
\begin{@twocolumnfalse}
\centering
\scriptsize
\setlength{\tabcolsep}{3pt}
\renewcommand{\arraystretch}{1.08}
\begin{tabular*}{\textwidth}{@{}>{\raggedright\arraybackslash}p{0.14\textwidth}@{\extracolsep{\fill}}>{\raggedright\arraybackslash}p{0.22\textwidth}>{\raggedright\arraybackslash}p{0.20\textwidth}>{\raggedright\arraybackslash}p{0.21\textwidth}>{\raggedright\arraybackslash}p{0.13\textwidth}@{}}
\toprule
\textbf{Leakage class} & \textbf{Rejected from retained artifacts} & \textbf{Allowed during current episode} & \textbf{Allowed retained guidance} & \textbf{Audit check} \\
\midrule
Task answer leakage & Final answers, exact patches, reference-skill text, target output values, or copied solution steps. & Task instructions, visible files, execution traces, and verifier feedback for the active repair. & Output-format checks, validation routines, and general execution order. & Reusable without the audited instance. \\
Instance identifiers & Hard-coded local paths, filenames, IDs, constants, or values not required as reusable behavior. & Task-declared paths, schemas, and assets when they are discovered or supplied by the current task. & Instructions to discover task-declared paths, schemas, and assets at runtime. & No frozen one-off literals. \\
Verifier leakage & Hidden tests, reward-script shortcuts, expected outputs, magic thresholds, or guessed oracle behavior. & Observable verifier contracts and failure messages exposed in the current trace. & Verifier-facing checks stated as format, tolerance, sentinel, or rerun procedures. & Does not infer hidden oracle facts. \\
Overfit repair & A procedure whose trigger or action only matches the audited instance. & Local diagnosis evidence that explains why the current skill failed. & A condition-action rule that transfers to sibling tasks with different files or values. & Trigger names the failure type, not the instance. \\
Principle leakage & Absorbed memory that copies task facts, answers, or benchmark-specific literals. & Calibration or revision evidence before filtering and abstraction. & A compact repair principle with an evidence-backed trigger, operation, and transfer guard. & Principle remains task-family level. \\
\bottomrule
\end{tabular*}
\refstepcounter{table}\label{tab:manual-leakage-audit}
\vspace{0.04in}
\noindent\begin{minipage}{0.98\textwidth}
\footnotesize Table~\thetable: Manual leakage audit checklist for retained skills, revision traces, and absorbed principles. The wide form separates current-episode evidence use from retained reusable guidance.
\end{minipage}
\vspace{0.10in}
\end{@twocolumnfalse}
]

\section{Manual Leakage Audit}
\label{app:manual-leakage-audit}

We audit evaluation-time skill leakage using the boundary summarized in Table~\ref{tab:manual-leakage-audit}: whether a retained skill or absorbed principle functions as reusable guidance rather than as a task solution \citep{li2026skillsbench}. This is distinct from pretraining contamination. \textsc{SkillRevise} is allowed to use task instructions, execution traces, and verifier feedback during the current revision episode, but the retained artifact should not encode task-instance answers, hidden verifier facts, or one-off shortcuts that would solve only the audited instance.

\subsection{Audit scope}
We manually inspected sampled retained artifacts across SkillsBench, SkillLearnBench-Random, SWE-Skills-Bench-Hard, and ALFWorld. The inspected units include selected revised skills, revision traces used to justify edits, and absorbed ALFWorld principles. Raw task logs and transient failed candidates may contain instance-local evidence because they are diagnosis inputs; the audit target is the content retained for later use, transfer, or principle retrieval.

\subsection{Benchmark-specific boundaries}
For SkillsBench and SkillLearnBench-Random, task instructions and visible environment files can be used as current-task context, but retained skills must phrase paths, schemas, and values as discovery or validation steps rather than as memorized constants. For SWE-Skills-Bench-Hard, repository prompts often name files, tests, or issue-specific behavior; using such information during the current repair is expected, while retaining a repository-specific patch recipe or reference-skill content as a general skill is treated as leakage. For ALFWorld, absorbed principles may describe repairs such as object tracking or admissible-action recovery, but should not memorize a room layout, object placement, or trajectory from a calibration task.

\subsection{Outcome and residual risk}
The audited retained artifacts did not reveal direct answer leakage. When revision traces contained task-specific observations, we kept them as provenance for the current candidate but required the retained skill or principle to rewrite them into reusable discovery, validation, fallback, or preservation rules. This manual audit cannot prove that every transient raw candidate was leakage-free, and verifier-visible feedback can still encourage overfitting to observable checks. We therefore treat the audit as a qualitative safeguard alongside protocol controls: no online absorption from evaluation tasks, verifier re-execution before selection, and conservative calibration-time filtering.

\section{Additional Limitations}
\label{app:additional-limitations}

\textbf{Verifier dependence.}
\textsc{SkillRevise} relies on verifier-visible execution evidence, whose limitations take two distinct forms. A flaky verifier can produce inconsistent outcomes for the same candidate because of nondeterministic tests, random seeds, external-service state, concurrency, timeouts, or evaluator variability. Partial validation may instead be stable while covering only a subset of the underlying requirements. Diagnosis, preservation constraints, and utility-gated best-observed selection can reduce unsupported or regressive updates, but they cannot correct errors shared by Diagnosis and Utility; under our reported terminal-on-success protocol, an accidental pass may also terminate candidate generation early. For flaky settings, a natural extension is to repeat evaluation of the same candidate before the stopping decision and aggregate utility with a stability penalty, such as mean utility minus a variance term. Repeated evaluation cannot recover requirements missing from a partial verifier; this case instead requires orthogonal holdout checks or complementary LLM/human judgment. Such judges should provide auxiliary evidence rather than serve as the sole acceptance oracle, with disagreement triggering abstention or independent validation. Our experiments use reproducible benchmark-provided verifiers and do not empirically evaluate these extensions.

\textbf{Revision cost and stopping.}
The method trades additional model calls, tool use, and wall-clock time for better skills. Although maxrev3 and terminal-on-success stopping limit this cost, some successful revisions still solve tasks through longer or more expensive procedures. A fixed budget is also imperfect: easy tasks may need no revision, while hard tasks may require a domain method that cannot be recovered within three attempts. Adaptive stopping, cheaper diagnosis, and earlier detection of misleading skills remain future priorities.

\textbf{Scope and generalization.}
Our evaluation covers multiple executors, three main verifier-based benchmarks, and domain-specific ALFWorld and BFCL-v4 studies, but it remains centered on tasks with explicit, reproducible success criteria. It also leaves realistic long-horizon agent benchmarks such as AgentVista and Tool Decathlon outside our scope~\citep{su2026agentvista,li2026tooldecathlon}. The results do not establish behavior in long-running deployments, safety-critical settings, adversarial tasks, or environments where the agent defines its own success criteria. In addition, maxrev3 does not fall back to no-skill execution, so skill conditioning can still hurt tasks that are better solved by direct instance-specific exploration. Future work should study larger skill libraries, dynamic routing, and deployment settings with changing tools, distributions, and user preferences.

\textbf{Memory maintenance.}
The present experiments evaluate fixed banks and calibration-then-freeze updates, not lifelong memory management. Continuous deployment would additionally require consolidation, conflict resolution, pruning, and stability checks across repeated updates. A practical extension is to collect validated failures in a separate calibration buffer, admit consolidated principles only after held-out utility, transfer, and non-interference checks, and deploy the updated bank read-only in the next evaluation or serving cycle.

\section{Prompt Templates}
\label{app:prompts}

This appendix reports the prompt and prompt-like interfaces used by the implementation. We use the same method-level authoring and revision templates across SkillsBench, SkillLearnBench-Random, and SWE-Skills-Bench-Hard; only the task-specific placeholder contents differ by benchmark and instance. Task-specific instructions, execution traces, and retrieved principles are represented as placeholders. Platform API configuration, credentials, and provider-specific hidden system prompts are omitted because they are not part of the proposed revision operator.

\caseheading{Skill authoring prompts.}
\begin{promptbox}[authorframe]{Initial Skill Authoring Prompt}
\textbf{Instruction.} You are writing a task-family skill for an LLM agent. Use platform-agnostic skill-authoring constraints: write a reusable procedure, keep only behavior-changing guidance, make workflow order and checkpoints explicit, validate inputs/files/tools/assumptions before acting, ground actions in the current environment, include fallback handling, add strict constraints, and avoid hard-coded paths, versions, or commands unless they are verified invariants.

\textbf{Inputs.} \texttt{[TASK\_FAMILY]}, \texttt{[SOURCE\_TASK]}, \texttt{[ACCEPTANCE\_CRITERIA]}, and optionally \texttt{[RETRIEVED\_PRINCIPLES]}.

\textbf{Principle use.} Use relevant principles as operating guidance, not as text to copy. The final skill must not mention principle IDs or copy task-specific answers, hidden verifier values, one-off constants, or unverified paths.

\textbf{Output.} Return only Markdown with sections \texttt{\# <Skill Name>}, \texttt{Purpose}, \texttt{When to Use}, \texttt{Procedure}, and \texttt{Constraints / Pitfalls}.
\end{promptbox}

\caseheading{Task execution prompts.}
\begin{promptbox}[execframe]{SkillsBench/Codex Task Execution Prompt}
\textbf{Instruction.} You are executing one SkillsBench task inside the provided workspace. Complete the task by editing or creating the required files in this workspace. Do not ask follow-up questions. Use local files and validation commands when available. When finished, provide a concise summary of what changed and what validation you ran.

\textbf{Optional skill guidance.} If \texttt{[INSTALLED\_SKILL]} is nonempty, use it as execution guidance. Treat it as reusable guidance, not as a task answer.

\textbf{Inputs.} \texttt{[TASK\_INSTRUCTION]} and optionally \texttt{[INSTALLED\_SKILL]}.

\textbf{Baseline use.} The no-skill baseline uses this execution prompt with an empty \texttt{[INSTALLED\_SKILL]}. Skill-Creator and \textsc{SkillRevise} use the same execution prompt after installing the generated or revised skill.
\end{promptbox}

\begin{promptbox}[execframe]{Benchmark-Native Execution Interface}
\textbf{Used for.} SkillLearnBench-Random and SWE-Skills-Bench-Hard run through their benchmark-native executor interfaces after export into the shared task-directory harness.

\textbf{Interface.} The executor receives a task directory containing \texttt{instruction.md}, environment files, tests or verifier scripts, and task metadata. When a skill is present, the harness materializes it as the benchmark skill source or skill directory; otherwise the same task runs with no installed skill.

\textbf{Prompt provenance.} The benchmark or agent framework supplies its normal task prompt from the task interface. We do not add a second method-specific task-solving prompt beyond installing the skill artifact and passing the same task instruction, environment, and verifier contract.
\end{promptbox}

\begin{promptbox}[execframe]{ALFWorld Interactive Action Prompt}
\textbf{Static context.} You are solving an ALFWorld text-only task. At each turn choose one valid environment action, preferably exactly one of the admissible commands. Explore only as needed, track object and receptacle names exactly, and stop after the environment reports success. The context includes \texttt{[TASK\_INSTRUCTION]}, \texttt{[PROBLEM\_PATH]}, and optionally reusable skill guidance.

\textbf{Per-turn inputs.} \texttt{[TURN\_INDEX]}, \texttt{[MAX\_TURNS]}, optional compact trajectory memory, \texttt{[CURRENT\_OBSERVATION]}, and \texttt{[ADMISSIBLE\_COMMANDS]}.

\textbf{Action policy.} Choose exactly one admissible command; do not repeat a failed action unless alternatives are exhausted. If actions alternate between the same locations, choose a new admissible destination or inspect a new relevant object. For look-at-object-in-light tasks, after taking the target object, actively search unvisited destinations for a lamp or DeskLamp.

\textbf{Output.} Return exactly one action string and no prose.
\end{promptbox}

\caseheading{Diagnosis prompts.}
\begin{promptbox}[diagframe]{LLM Diagnosis Prompt}
\textbf{Instruction.} Diagnose why an LLM-authored skill helps, hurts, or fails to help a task execution. Use paired execution evidence, not stylistic preference alone.

\textbf{Inputs.} \texttt{[TASK\_ID]}, \texttt{[TASK\_FAMILY]}, \texttt{[TASK\_INSTRUCTION]}, \texttt{[ACCEPTANCE\_CRITERIA]}, \texttt{[CURRENT\_SKILL]}, authoring-prior violations, \texttt{[NO\_SKILL\_TRACE]}, \texttt{[WITH\_SKILL\_TRACE]}, and \texttt{[UTILITY]}.

\textbf{Allowed labels.} path, schema, method, environment, incomplete execution, data understanding, constraint violation, verification gap, efficiency timeout, overfit or hardcode, context pollution, false certainty, wrong abstraction level, and related skill-defect labels used by the harness.

\textbf{Output.} Return only JSON with keys \texttt{labels}, \texttt{evidence}, \texttt{causal\_judgment}, \texttt{rewrite\_targets}, and \texttt{summary}. Evidence items contain \texttt{source}, \texttt{snippet}, and \texttt{reason}. Non-LLM diagnosis runs construct the same fields deterministically from traces and verifier outcomes.
\end{promptbox}

\caseheading{Skill revision prompts.}
\begin{promptbox}[revisionframe]{Structured Revision Prompt}
\textbf{Instruction.} Revise this task-family skill using execution evidence and Principle Memory. The revised skill should teach a reusable non-regression workflow, not memorize this task's answer. Use selected principles as repair operators, not as extra prose pasted into the skill.

\textbf{Inputs.} \texttt{[PREVIOUS\_REVISION\_TRACE]}, \texttt{[TASK\_FAMILY]}, \texttt{[TASK\_INSTRUCTION]}, \texttt{[ACCEPTANCE\_CRITERIA]}, \texttt{[CURRENT\_SKILL]}, \texttt{[DIAGNOSIS\_LABELS]}, \texttt{[CAUSAL\_JUDGMENT]}, \texttt{[EVIDENCE]}, \texttt{[REWRITE\_TARGETS]}, and top-$k$ \texttt{[RETRIEVED\_PRINCIPLES]}.

\textbf{Protocol.} Build a verifier contract, failure ledger, and preserve ledger before editing. Select at most three retrieved principles, and for each selected principle state why it applies, which failed check/evidence it addresses, what skill operation it induces, and what preserve constraint it protects. If the same failure repeats across attempts, escalate from local patching to method-level repair. Every revision must introduce or preserve an execution anchor with action, expected evidence, and placement. Edit only the responsible trigger, procedure, or constraint unless evidence implicates shared setup. Remove task-instance answers, brittle literals, hidden-verifier guesses, and instructions that only work for the observed instance.

\textbf{Output.} Return exactly two blocks: first \texttt{REVISION\_TRACE\_JSON}, then \texttt{REVISED\_SKILL\_MARKDOWN}.
\end{promptbox}

\begin{promptbox}[schemaframe]{\texttt{REVISION\_TRACE\_JSON} Schema}
The structured revision prompt requires a machine-readable trace before the revised skill. The required fields are:
\begin{itemize}[leftmargin=*,nosep]
\item \texttt{verifier\_contract}: observable requirements only, including required output paths, schemas or formats, numeric thresholds, and pass/fail assertions. Unknown requirements must remain unknown rather than guessed.
\item \texttt{failure\_ledger}: failed checks, actual behavior, likely cause, and a lightweight failure type such as path, schema, method, environment, incomplete execution, data understanding, constraint violation, verification gap, efficiency timeout, overfit or hardcode, or unknown.
\item \texttt{preserve\_ledger}: passed checks and successful choices to keep. Preserved behavior should not be removed, renamed, contradicted, or replaced unless it directly conflicts with the failed check.
\item \texttt{selected\_principles}: for each selected principle, principle ID, reason for selection, addressed failure, induced skill operation, and preserve constraint.
\item \texttt{ignored\_principles}: retrieved principles not used, with evidence-grounded reasons.
\item \texttt{repeated\_failure\_escalation}: whether repeated failure triggered a higher-level repair and what escalation action was taken.
\item \texttt{execution\_anchors}: concrete action, expected observable evidence, and placement in the revised skill.
\item \texttt{acceptance\_signals}: expected utility improvement, expected failed assertions reduced, and preserve risk.
\end{itemize}
The second block must be the revised skill Markdown with the same standard sections used by the initial authoring prompt.
\end{promptbox}

\caseheading{Ablation prompts.}
\begin{promptbox}[ablationframe]{Free-form Revision Prompt}
\textbf{Instruction.} Improve this LLM-authored task skill using the observed execution feedback. Do not use any predefined failure taxonomy, repair-principle checklist, or structured defect labels. Keep useful parts of the current skill, remove misleading parts, and make the revised skill reusable. Do not regress verifier checks that already passed while addressing the observed feedback.

\textbf{Inputs.} \texttt{[TASK\_FAMILY]}, \texttt{[TASK\_INSTRUCTION]}, \texttt{[ACCEPTANCE\_CRITERIA]}, \texttt{[CURRENT\_SKILL]}, and \texttt{[OBSERVED\_EXECUTION\_FEEDBACK]}.

\textbf{Output.} Return only the revised skill Markdown with sections \texttt{\# <Skill Name>}, \texttt{Purpose}, \texttt{When to Use}, \texttt{Procedure}, and \texttt{Constraints / Pitfalls}.
\end{promptbox}

\begin{promptbox}[ablationframe]{Unstructured Repeated Revision Prompt}
\textbf{Instruction.} Revise the current task-family skill so that it performs better using the task description and the leakage-safe execution/verifier feedback below.

\textbf{Inputs.} \texttt{[TASK\_INSTRUCTION]}, \texttt{[ACCEPTANCE\_CRITERIA]}, \texttt{[CURRENT\_SKILL]}, and \texttt{[LEAKAGE\_SAFE\_EXECUTION/VERIFIER\_FEEDBACK]}.

\textbf{Output.} Return only the revised skill Markdown using the same standard sections as the initial skill.

\textbf{Protocol.} Starting from the matched $S_0$, the prompt is applied for three rewrite--re-execute rounds under maxrev3. The rewriter receives no structured Diagnosis, Principle Memory, preserve ledger, or Execution Anchors; the reported result is measured after the third round.
\end{promptbox}

\begin{tcolorbox}[
  enhanced jigsaw,
  colback=promptback,
  colframe=schemaframe,
  colbacktitle=schemaframe,
  coltitle=white,
  fonttitle=\bfseries\footnotesize,
  title={LLM Backend Wrapper Prompt},
  boxrule=0.8pt,
  arc=2mm,
  left=5pt,
  right=5pt,
  top=4pt,
  bottom=4pt,
  before skip=2pt,
  after skip=2pt,
  fontupper=\scriptsize,
  before upper={\raggedright\setlength{\parindent}{0pt}}]
\textbf{Instruction.} You are the LLM backend for \textsc{SkillRevise}; follow the requested output format exactly and add no unrelated commentary.

\textbf{Purpose field.} The wrapper receives a purpose such as \texttt{skill\_authoring}, \texttt{skill\_revision}, \texttt{skill\_revision\_freeform}, or \texttt{skill\_diagnosis}, then passes the corresponding method prompt as \texttt{[PROMPT]}.

\textbf{Command-line LLM variant.} Return only the final answer; do not include progress notes or run tools unless necessary.
\end{tcolorbox}

\section{Principle Memory Entries}
\label{app:principle-memory-entries}
\small

The seed Principle Memory contains reusable repair operators rather than task answers. Each entry has a trigger, repair operation, and guard against over-specific retention.

\begin{tcolorbox}[
  enhanced jigsaw,
  breakable,
  colback=promptback,
  colframe=promptframe,
  boxrule=0.6pt,
  arc=1mm,
  left=5pt,
  right=5pt,
  top=4pt,
  bottom=4pt,
  before skip=2pt,
  after skip=4pt,
  fontupper=\scriptsize]
\principlecard{Workflow checkpointing}{broad advice without verifiable intermediate decisions.}{rewrite the procedure as ordered checkpoints: discover, validate, act, verify, and recover.}{encode the check, not a one-task answer.}\par\medskip
\principlecard{Input schema validation}{malformed fields, wrong types, missing keys, or numeric mismatches.}{reload produced artifacts and assert required keys, types, shapes, ranges, and serialization formats.}{check task-declared schema, not expected values.}\par\medskip
\principlecard{Environment-output grounding}{expected artifact is missing or written outside the checked path.}{check task-specified output paths and inspect mounts, permissions, scripts, or supported write routes if needed.}{use exact paths only when supplied or locally discovered.}\par\medskip
\principlecard{Verifier-contract alignment}{plausible domain logic misses a verifier convention.}{restate sentinels, formats, tolerances, traversal rules, or scoring rules as pre-finalization checks.}{record the convention class, not hidden answers.}\par\medskip
\principlecard{Fallback after tool failure}{a command, tool, file, endpoint, or assumption fails without a valid alternative.}{add one bounded fallback branch, choose the closest supported alternative, and rerun the smallest check.}{keep fallback short and task-directed.}\par\medskip
\principlecard{Transfer-preserving repair}{repair targets one instance, path, literal, or answer rather than reusable behavior.}{replace instance-specific content with a trigger condition and reusable decision rule.}{require usefulness for sibling tasks with different files or values.}\par\medskip
\principlecard{Trigger noninterference}{skill guidance adds cost or steers execution when it is not needed.}{narrow \emph{When to Use} and add exclusions so the skill fires only when it changes an execution decision.}{prefer concise trigger boundaries over broad domain labels.}
\end{tcolorbox}

\subsection{Principle-Guided Repair Contrasts}
\label{app:principle-memory-examples}

We analyze five principle-guided contrasts involving the seed principles in Appendix~\ref{app:principle-memory-entries}. In each contrast, the standard run selects a revised skill containing the principle and succeeds, while the corresponding w/o Principles ablation fails. The boxes below pair the weak or failed skill content with the corresponding principle-guided revision. We omit task answers, raw verifier values, hidden expected outputs, and instance-specific assertions.

\caseheading{Input schema validation.}
\begin{promptbox}[ablationframe]{Weak skill: schema-shaped output}
Extract the handbook rules, write the requested JSON artifacts, and run the provided checks. Treat a parseable file with the expected top-level fields as ready unless a command explicitly fails.
\end{promptbox}
\begin{promptbox}[authorframe]{Revised skill: reload and recompute}
Write artifacts atomically; reload each produced JSON file; check required keys, types, ordering, and rounding; then independently recompute the task-critical metric from the task-visible inputs before finalizing.
\end{promptbox}
\noindent\textbf{Analysis.} In the manufacturing maintenance contrast, the standard run selects $v_1$ and succeeds, while w/o Principles selects $v_0$ and later no-principle candidates continue to fail on a task-critical numeric artifact. The weak skill validates that the artifact looks like the requested output, but it does not prove that the computation inside the artifact is correct. The revised skill changes the failure boundary: a syntactically valid but semantically wrong file becomes a blocked pre-finalization state, forcing the executor to repair the computation method instead of merely producing well-formed JSON.

\caseheading{Fallback after tool failure.}
\begin{promptbox}[ablationframe]{Weak skill: continue after missing evidence}
Search for the required itinerary entities, record the tools used, and continue with the best available plan when one source or route lookup is unavailable.
\end{promptbox}
\begin{promptbox}[authorframe]{Revised skill: bounded fallback}
Maintain a required-evidence ledger. If a route or source is missing, inspect available tools, choose a verified equivalent, rerun only the affected search, and preserve already-passing itinerary constraints.
\end{promptbox}
\noindent\textbf{Analysis.} In the travel planning contrast, the standard run selects $v_3$ and succeeds, while w/o Principles selects $v_1$ and fails despite containing a generic checklist for searched itinerary construction. The weak skill treats missing evidence as something to work around with a plausible plan. The revised skill instead treats it as a local recovery problem. This works because the repair is neither a full replanning pass nor a guess: it is a bounded fallback that keeps the valid parts of the itinerary stable while recovering the specific missing evidence.

\caseheading{Transfer-preserving repair.}
\begin{promptbox}[ablationframe]{Weak skill: instance-shaped patch}
After observing a failure, patch the immediate case with concrete local filenames, literals, or one-task assumptions that explain the observed mismatch.
\end{promptbox}
\begin{promptbox}[authorframe]{Revised skill: reusable repair rule}
Discover paths and schemas from the current task; preserve handbook-grounded formulas, sorting, rounding, and schema checks; avoid storing instance identifiers, expected values, or one-task constants.
\end{promptbox}
\noindent\textbf{Analysis.} The same manufacturing maintenance contrast shows why transfer preservation matters. The standard $v_1$ succeeds, whereas w/o Principles stays at $v_0$ as the selected skill and later no-principle revisions keep failing on the same method-level defect. The weak repair could solve the observed failure by memorizing the instance. The principle-guided repair instead expresses the fix as a task-family procedure: discover the local contract, apply the reusable computation rule, and validate the emitted artifact. That is why the revised skill remains a skill rather than an answer patch.

\caseheading{Workflow checkpointing.}
\begin{promptbox}[ablationframe]{Weak skill: solve then report}
Parse the network, solve the dispatch problem, write the report, and run a final check after the report has been produced.
\end{promptbox}
\begin{promptbox}[authorframe]{Revised skill: checkpointed workflow}
Run discover, solve, pre-write validate, write, reload, and post-write verify as separate checkpoints. Recompute report totals, loading, reserve, margin, and cost from the emitted artifact.
\end{promptbox}
\noindent\textbf{Analysis.} For the grid dispatch contrast, the standard run selects $v_2$ and succeeds, while w/o Principles selects $v_0$ and its later candidate still fails after producing a report that is not economically consistent. The weak skill contains the right domain ingredients, but it does not connect the solver state to the serialized artifact. Checkpointing makes the final report itself the object that must satisfy the contract, so objective drift or serialization drift is caught even when earlier feasibility checks look plausible.

\caseheading{Trigger noninterference.}
\begin{promptbox}[ablationframe]{Weak skill: broad domain trigger}
Use grid-dispatch guidance whenever the task appears related to power systems, and include general domain reasoning before producing the final report.
\end{promptbox}
\begin{promptbox}[authorframe]{Revised skill: narrow execution trigger}
Trigger only for structured DC-dispatch report tasks. Keep contract discovery, optimization, serialized-report checks, and bounded recovery; exclude unrelated grid analysis or narrative-only tasks.
\end{promptbox}
\noindent\textbf{Analysis.} Grid dispatch also illustrates how trigger scope affects repair. The no-principle selected skill is broad enough to guide a dispatch attempt, but it does not focus the executor on the final report gate. The principle-guided revision removes low-signal domain prose and narrows the trigger to the exact class of tasks where the skill changes an execution decision. This reduces skill-induced distraction and lets the executor spend its budget on the checks that determine success.

Across these contrasts, principles convert verifier feedback into reusable repair operations rather than task answers, helping the standard run recover failures that the w/o Principles ablation continues to rewrite around.

\normalsize

\begin{figure*}[t]
  \refstepcounter{section}
  \noindent{\large\bfseries \thesection\quad GPT-5.5 Per-Task Heatmap\par}
  \vspace{0.4em}
  \centering
  \includegraphics[
    height=0.82\textheight,
    width=0.95\textwidth,
    keepaspectratio
  ]{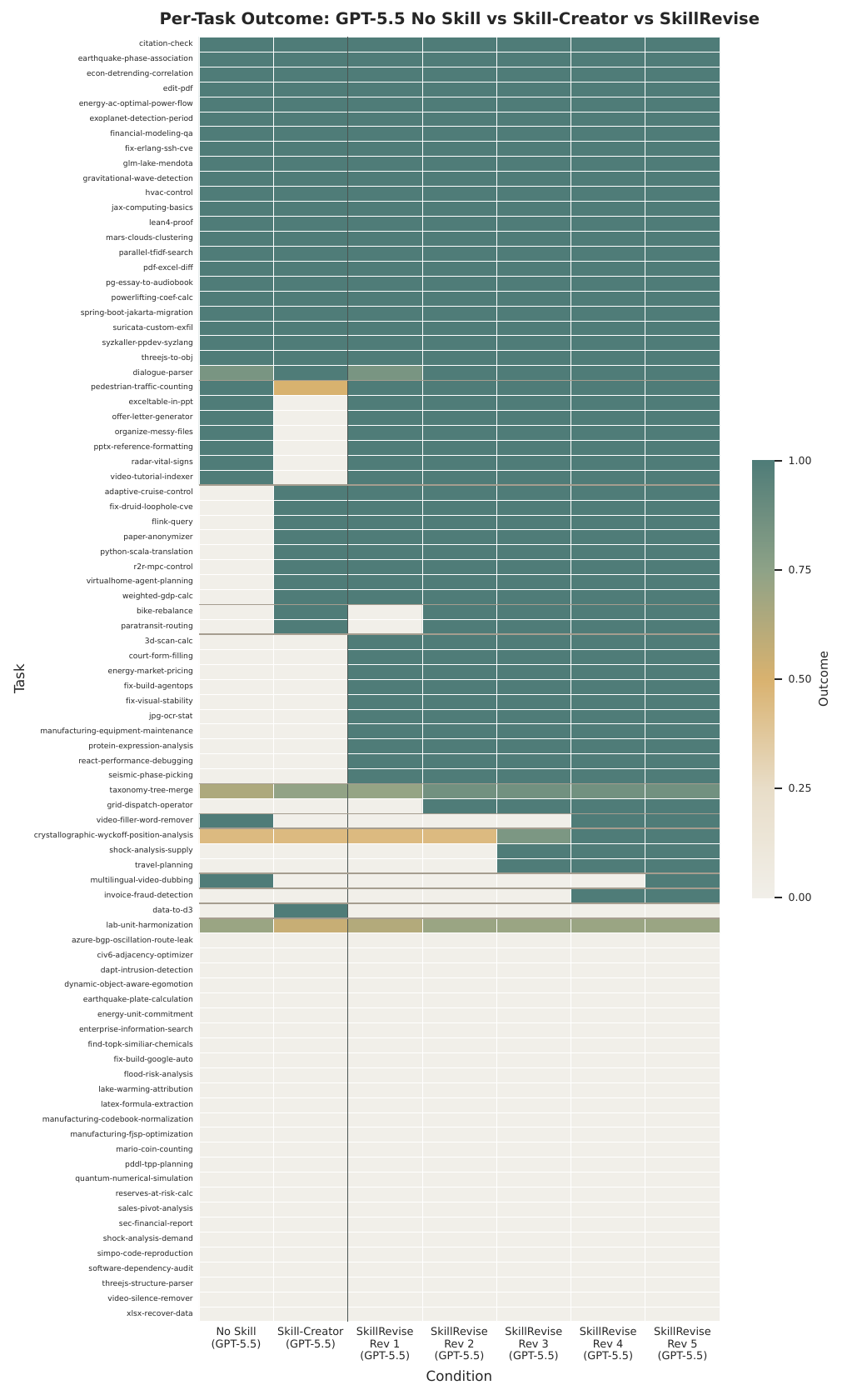}
  \caption{
    Per-task verifier outcome heatmap for GPT-5.5 across methods on SkillsBench.
    Rows are grouped by revision-response pattern, placing tasks solved by all methods before tasks newly solved by revision, partially improved tasks, and persistently difficult tasks.
    Columns compare the no-skill baseline, Skill-Creator, and revised skills under revision budgets Rev1--Rev5.
    Deep-teal cells indicate higher verifier outcome scores, gold cells indicate partial scores, warm-gray cells indicate low-scoring or failed executions, and thin separators mark task groups.
  }
  \label{fig:gpt55-task-heatmap}
\end{figure*}

\begin{figure*}[t]
  \refstepcounter{section}
  \noindent{\large\bfseries \thesection\quad Opus-4.7 SkillLearnBench Per-Task Heatmap\par}
  \vspace{0.4em}
  \centering
  \includegraphics[
    height=0.82\textheight,
    width=0.95\textwidth,
    keepaspectratio
  ]{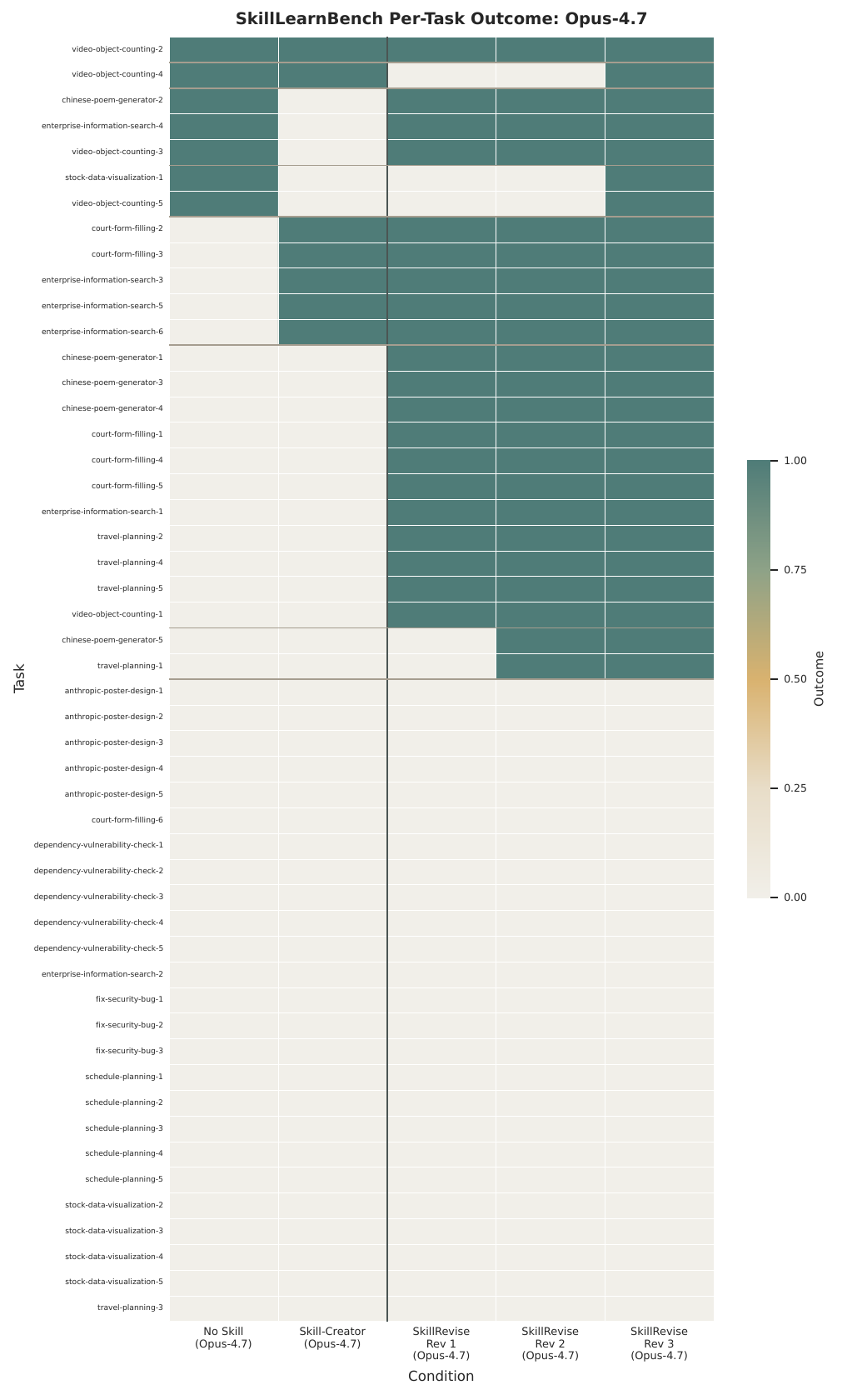}
  \caption{
    Per-task verifier outcome heatmap for Opus-4.7 across methods on SkillLearnBench-Random.
    Columns compare the no-skill baseline, Skill-Creator, and budgeted \textsc{SkillRevise} selections after Rev1, Rev2, and Rev3 reconstructed from the recorded revision history.
    Rows cover the full 50-task selected suite.
    Rows are sorted to concentrate revision-side successes in the upper-right region while leaving persistent failures at the bottom.
  }
  \label{fig:opus47-skilllearnbench-task-heatmap}
\end{figure*}

\begin{figure*}[t]
  \refstepcounter{section}
  \noindent{\large\bfseries \thesection\quad Qwen-3.6-Plus SWE-Skills-Bench Per-Task Heatmap\par}
  \vspace{0.4em}
  \centering
  \includegraphics[
    height=0.82\textheight,
    width=0.95\textwidth,
    keepaspectratio
  ]{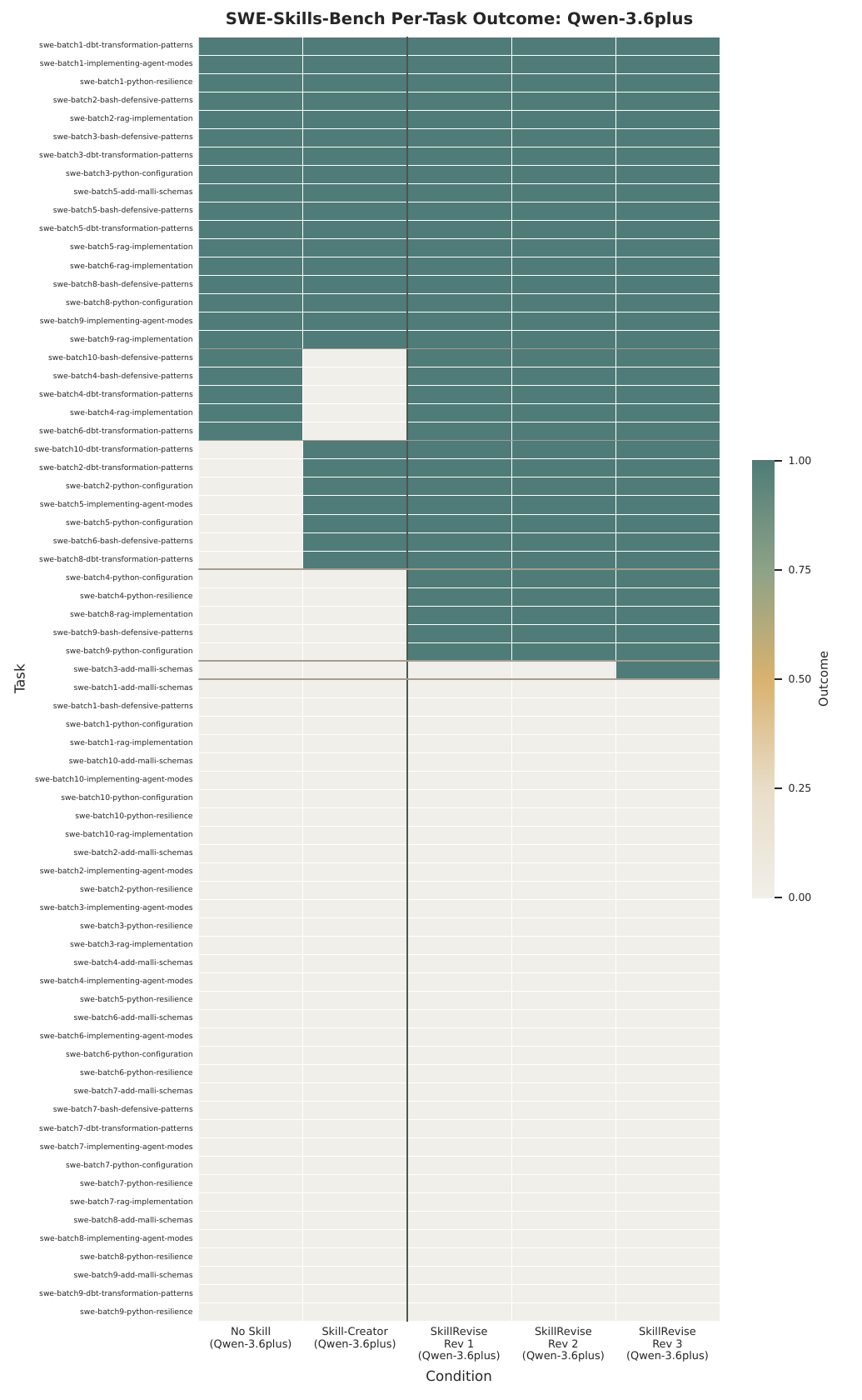}
  \caption{
    Per-task verifier outcome heatmap for Qwen-3.6-Plus across methods on SWE-Skills-Bench-Hard.
    Columns compare the no-skill baseline, Skill-Creator, and budgeted \textsc{SkillRevise} selections after Rev1, Rev2, and Rev3 reconstructed from the recorded revision history.
    Rows are sorted by the binary response pattern so shared successes, revision-only successes, and persistent failures are visually grouped.
  }
  \label{fig:qwen36plus-swe-task-heatmap}
\end{figure*}

\end{document}